\documentclass{article}

\usepackage{microtype}
\usepackage{graphicx}
\usepackage{subfigure}
\usepackage{booktabs} 

\usepackage{amsmath,amssymb}
\usepackage{courier}

\usepackage[dvipsnames]{xcolor}

\usepackage{hyperref}
\usepackage{xr}
\usepackage[accepted]{icml2019}

\icmltitlerunning{Scalable Nonparametric Sampling from Multimodal Posteriors with the Posterior Bootstrap}

\DeclareMathOperator{\EX}{\mathbb{E}}
\DeclareMathOperator*{\argmax}{arg\,max}
\DeclareMathOperator*{\argmin}{arg\,min}
\DeclareMathOperator*{\localargmin}{local\, arg\,min}
\begin{document}

\twocolumn[
\icmltitle{Scalable Nonparametric Sampling from Multimodal Posteriors \\ with the Posterior Bootstrap}

\begin{icmlauthorlist}
\icmlauthor{Edwin Fong}{ox,ati}
\icmlauthor{Simon Lyddon}{ox}
\icmlauthor{Chris Holmes}{ox,ati}

\end{icmlauthorlist}

\icmlaffiliation{ox}{Department of Statistics, University of Oxford, Oxford, United Kingdom}
\icmlaffiliation{ati}{The Alan Turing Institute, London, United Kingdom}

\icmlcorrespondingauthor{Edwin Fong}{\mbox{edwin.fong@stats.ox.ac.uk}}

\icmlkeywords{Computational Statistics, Machine Learning, ICML}

\vskip 0.3in
]

\printAffiliationsAndNotice{} 
\hyphenpenalty=700
\exhyphenpenalty=700
\begin{abstract}

Increasingly complex datasets pose a number of challenges for Bayesian inference. Conventional posterior sampling based on Markov chain Monte Carlo can be too computationally intensive, is serial in nature and mixes poorly between posterior modes. Furthermore, all models are misspecified, which brings into question the validity of the conventional Bayesian update. We present a scalable Bayesian nonparametric learning routine that enables posterior sampling through the optimization of suitably randomized objective functions. A Dirichlet process prior on the unknown data distribution accounts for model misspecification, and admits an embarrassingly parallel posterior bootstrap algorithm that generates independent and exact samples from the nonparametric posterior distribution. Our method is particularly adept at sampling from multimodal posterior distributions via a random restart mechanism, and we demonstrate this on Gaussian mixture model and sparse logistic regression examples.
\end{abstract}
\hyphenpenalty=100
\exhyphenpenalty=100

\section{Introduction}
\label{intro}
As datasets grow in complexity and size, Bayesian inference becomes increasingly difficult. The posterior is often intractable, so we resort to simulation methods for inference via Markov chain Monte Carlo (MCMC), which is inherently serial and often too computationally expensive in datasets with a large number of data points \cite{Bardenet:2017}. MCMC further struggles with multimodal posteriors which arise in many settings including mixture models \cite{Jasra2005} or non-convex priors \cite{Seeger}, as the MCMC sampler can become trapped in local modes  \cite{Rudoy2006}. Current methods to sample from multimodal posteriors with MCMC include parallel tempering \cite{Neal1996} and adaptive MCMC \cite{Pompe2018}, but the associated computational cost is high. Posterior approximation with variational Bayes (VB) \cite{Blei2017} is a faster alternative, but it is generally difficult to quantify the quality of the approximation, and is thus problematic if accurate uncertainty quantification is desired  \cite{Giordano2015}.

A further methodological issue facing Bayesian inference is the fact that all models are false. The increasing scale of datasets exacerbates the effects of model misspecification \cite{Walker2013}, as the true sampling distribution is meaningfully different from the parametric family of distributions of the model. There is rarely formal acknowledgement of model misspecification which can lead to inconsistencies \cite{watson2016,Grunwald2017}.

Bayesian nonparametric learning (NPL) introduced by \citet{Lyddon} allows for the use of statistical models without assuming the model is true. NPL uses a nonparametric prior centred on a parametric model, and returns a nonparametric posterior over the parameter of interest. The method focuses on accounting for model misspecification and for posterior approximation such as from Variational Bayes (VB) by placing a mixture of Dirichlet processes \cite{Antoniak1974} prior on the sampling distribution. In addition to the acknowledgement of model misspecification, the method admits an embarrassingly parallel Monte Carlo sampling scheme consisting of randomized maximizations. However, in most cases this method requires sampling the Bayesian posterior, which is  computationally expensive  for complex models.

\subsection{Our Contribution}

In this work, we propose a simplified variant of NPL that utilises a Dirichlet process (DP) prior on $F_0$ instead of a mixture of Dirichlet processes (MDP) prior. This allows us to perform inference directly and detaches the nonparametric prior from the prior of the model parameter of interest. Instead of centering on a Bayesian posterior, we center the DP on a sampling distribution which encapsulates our prior beliefs.  This simpler choice of prior also has desirable theoretical properties and is highly scalable as we no longer need to sample from the Bayesian posterior.  Our method can handle a variety of statistical models through the choice of the loss functions, and can be applied to a wide range of machine learning settings as we will demonstrate in Section \ref{examples}. Our method implies a natural noninformative prior, which may be relevant when the number of data points is substantially larger than the number of parameters.

The posterior bootstrap sampling scheme was introduced by \citet{Lyddon} under the NPL framework, and we inherit its computational strengths such as parallelism and exact inference under a Bayesian nonparametric model. Independent samples from the nonparametric posterior are obtained through the optimization of randomized objective functions, and we obtain the weighted likelihood bootstrap \cite{Newton1994} as a special case. Furthermore, sampling from multimodal posteriors  now involves a non-convex optimization at each bootstrap sample that we solve through local search and random restart. We demonstrate that our method recovers posterior multimodality on a Gaussian Mixture Model (GMM) problem. We further show that our method is computationally much faster than conventional Bayesian inference with MCMC, and has superior predictive performance  on real sparse classification problems. Finally, we utilize the computational speed of NPL to carry out a Bayesian sparsity-path-analysis for variable selection on a genetic dataset.

\section{Bayesian Nonparametric Learning}
Assume that we have observed $y_{1:n} \overset{iid}{\sim} F_0$, where $y_{1:n}$ is a sequence of $n$ i.i.d. observables and $F_0$ is the unknown sampling distribution. We may be interested in a parameter $\theta \in \Theta \subseteq \mathbb{R}^p $, which indexes a family of probability densities $\mathcal{F}_\Theta = \{ f_\theta(y) ; \theta \in \Theta \}$.  Conventional Bayesian updating of the prior to the posterior via Bayes' theorem formally assumes that $F_0$ belongs to the model $F_\Theta$, which is questionable in the presence of complex and large datasets. This assumption is not necessary for NPL. We derive the foundations of NPL by treating parameters as functionals of $F_0$, with model fitting as a special case. 
\subsection{The Parameter of Interest}
We define our parameter of interest as
\begin{equation} \label{eq1}
\theta_0(F_0) = \argmin_{\theta} \int l(y,\theta) dF_0(y)
\end{equation}
where $l(y,\theta)$ is a loss function, and its form can be used to target statistics of interest. For example, setting  $l(y,\theta) = |y-\theta|$ returns the median and $ (y-\theta)^2$ returns the mean.

The loss function of particular interest is $l(y,\theta) = -\log f_\theta(y)$, where  $f_\theta$ is the density of some parametric model. The value of $\theta_0$ minimises the Kullback-Leibler divergence $\text{KL}(f_0||f_\theta)$, which is the parameter of interest in conventional Bayesian analysis \cite{Walker2013,Bissiri2016}. We have not assumed that $F_\Theta$ contains $F_0$, and $\theta_0$ in this case does not have any particular generative meaning as it is simply the parameter that satisfies (\ref{eq1}). 

\subsection{The Dirichlet Process Prior}
As the sampling distribution is unknown, we place a DP prior on $F_0$
\begin{equation}\label{eq2}
[F | \alpha,F_\pi ]\sim \text{DP}\left(\alpha, F_\pi \right)
\end{equation}
where $F_\pi$ is our prior centering measure, and $\alpha$ is the strength of our belief. 
 \vspace{-0.1in}
\paragraph{The base measure $\boldsymbol{F_\pi}$}
We encode our prior knowledge about the sampling distribution in the measure $F_\pi$. If we believe a particular model $f_\theta$ to be accurate, and have prior beliefs about $\theta$ encoded in $\pi(\theta)$, a sensible choice for the density of $F_\pi$ is $f_\pi(y) = \int f_\theta(y) d\pi(\theta)$. Alternatively, we could directly specify $f_\pi$ as a density that accurately represents our beliefs without the burden of defining a joint distribution on $(y,\theta)$. In the presence of historical data $\hat{y}_{1:\hat{n}}$, a suitable choice for $F_\pi$ is the empirical distribution of the historical data, i.e. $F_\pi (y) = \frac{1}{\hat{n}} \sum_{i=1}^{\hat{n}}\delta_{\hat{y}_i}(y)$ where $\delta$ is the Dirac measure. This is in a similar fashion to power priors \cite{ibrahim2000}. Further intuition is provided in Section \ref{prior_intuit} of the Supplementary Material.

It should be noted that we cannot directly include a prior on the parameter of interest $\theta_0$, only implicitly through $(\alpha, F_\pi )$.  Our prior is selected independently of the model of interest, and  this is appropriate under a misspecified model setting since we do not believe there to be a true $f_\theta$.  As all parameters of interest are defined as a functional of $F_0$ as in (\ref{eq1}), any informative prior on $F_0$ is thus informative of $\theta_0$. 
\vspace{-0.05in}
\paragraph{The concentration $\boldsymbol{\alpha}$ }
The size of $\alpha$ measures the concentration of the DP about $F_\pi$, and a large value corresponds to a smaller variance in a functional of the DP. We see in (\ref{eq3}) that the DP posterior base measure is a weighted sum of the prior $F_\pi$ and the empirical distribution $F_n = \frac{1}{n}\sum_{i=1}^n \delta_{y_i}$, with the weights proportional to $\alpha$ and $n$ respectively. We can thus interpret $\alpha$ as the effective sample size from the prior $F_\pi$.  One method of selecting $\alpha$ is through simulation of the prior distribution of $\theta$ via (\ref{eq1}) and tuning its variance. Alternatively, we can select $\alpha$ through the a priori variance of the mean functional (see Section \ref{mean_func} of the Supplementary Material). The special case of $\alpha = 0$ corresponds to the Bayesian bootstrap \cite{Rubin1981}, which in our case corresponds to a natural way to define an noninformative prior about $F_0$ (see \citet{gelman2013bayesian} for a review on noninformative priors). For  $n \gg p$, it may be suitable to set $\alpha = 0$ as the prior should have little influence and the Bayesian bootstrap is more computationally efficient. 

\subsection{The NPL Posterior }
From the conjugacy of the DP, the posterior of $F$ is
\begin{equation} \label{eq3}
\begin{aligned}
\left[F|y_{1:n}\right] \sim \text{DP}\left(\alpha +n, G_n\right),\\ G_n = \frac{\alpha}{\alpha+n}F_\pi + \frac{1}{\alpha+ n}\sum_{i=1}^n \delta_{y_i}.
\end{aligned}
\end{equation}
Our NPL posterior $\tilde{\pi}(\theta|y_{1:n})$ is thus
\begin{equation}\label{eq4}
\begin{aligned}
\tilde{\pi}(\theta|y_{1:n})  &= \int \pi(\theta |F) d\pi(F|y_{1:n}) 
\end{aligned}
\end{equation}
where $\pi(\theta|F) = \delta_{\theta_0(F)}(\theta)$; the delta arises as $\theta$ is a deterministic functional of $F$ as in (\ref{eq1}).  Properties of the NPL posterior follow from properties of the DP, e.g. draws of $F|y_{1:n}$ are almost surely discrete, so (\ref{eq1}) simplifies to
 \vspace{-0.05in}
\begin{equation} \label{eq5}
\theta(F) = \argmin_{\theta} \sum_{k=1}^\infty w_k l(\tilde{y}_k,\theta)
\end{equation}
where $w_{1:\infty} \sim \text{GEM}(\alpha+n)$  and $\tilde{y}_{1:\infty} \overset{iid}{\sim} G_n$ from the stick-breaking construction \cite{Sethuraman1994}. Formally, the GEM distribution is defined
\vspace{-0.05in}
\begin{equation}
v_k \sim \text{Beta}(1,\alpha + n), \quad w_k = v_k \prod_{j=1}^{k-1}(1-v_j).
\end{equation}
We preserve the theoretical advantages from \citet{Lyddon} due to the symmetries in the limits of the DP and the MDP for $\alpha \to 0$ and $n \to \infty$, where $\alpha$ also denotes the concentration parameter of the MDP. 
\paragraph{Consistency}
Under regularity conditions, the NPL posterior is consistent at $\theta_0$ as defined in (\ref{eq1}), from the properties of the DP (see \citet{vdV00,ghosal_2010,ghosal_vandervaart_2017} for details).  Interestingly, this is true regardless of the choice of $F_\pi$ and its support.  This is not the case in conventional Bayesian inference through Bayes' rule where the support of the prior must contain $\theta_0$ for posterior consistency. This is particularly reassuring in our misspecified model setting, as inferences about $\theta_0$ are robust to choices of $F_\pi$. 

\paragraph{Asymptotic dominance}
The NPL posterior predictive $\tilde{\pi}(\cdot|y_{1:n})$ for $\alpha = 0$ \textit{asymptotically dominates} the conventional Bayesian posterior predictive  $\pi(\cdot|y_{1:n})$  up to $o(n^{-1})$ under regularity conditions, i.e.
\begin{equation}
\begin{split}
\EX_{y_{1:n}\sim q}\left[ \text{KL}(q(\cdot)||\pi(\cdot | y_{1:n}))-\text{KL}(q(\cdot)||\tilde{\pi}(\cdot | y_{1:n})) \right] \\= K(q(\cdot)) + o(n^{-1})
\end{split}
\end{equation}
for all distributions $q$, where $K$ is a non-negative and possibly positive real-valued functional. This states that compared to the Bayesian posterior predictive, the NPL posterior predictive is closer  in expected KL divergence to the true $F_0$ up to $o(n^{-1})$. The proof for the MDP case is given in Theorem 1 of \citet{Lyddon}, and the above follows from the equivalence of the MDP and the DP for $\alpha = 0$. 

\subsection{Sampling from the NPL Posterior}\label{samp}

In almost all cases, $\tilde{\pi}(\theta|y_{1:n})$ is not tractable, but lends itself to a parallelizable Monte Carlo sampling scheme. It may be more intuitive to think of sampling $F$ from the posterior DP, then calculating (\ref{eq1}) to generate the sample from $\tilde{\pi}(\theta|y_{1:n})$, as shown in Algorithm \ref{NPLsamp}.
\begin{algorithm}[H]
 \caption{NPL Posterior Sampling}
 \label{NPLsamp}
     \begin{algorithmic}
       \FOR{$i=1$ {\bfseries to} $B$}
       \STATE{Draw $F^{(i)} \sim \text{DP} (\alpha+n, G_n) $}
       \STATE{$\theta^{(i)} = \argmin_{\theta} \int l(y,\theta) dF^{(i)}(y)$}
       \ENDFOR
    \end{algorithmic}
\end{algorithm}
\vskip -0.1in
Here $B$ is the number of posterior bootstrap samples. One advantage of this sampling scheme is that it is embarrassingly parallel as each of the $B$ samples can be drawn independently. We can thus take advantage of increasingly available multi-core computing, unlike in conventional Bayesian inference as MCMC is inherently sequential. 
 
\subsubsection{The Posterior Bootstrap}
Sampling from the DP exactly requires infinite computation time if $F_\pi$ is continuous, but approximate samples can be generated by truncation of the sum in (\ref{eq5}). For example, we could truncate the stick-breaking and set the remaining weights to 0. Alternatively, we could approximate $w_{1:T} \sim \text{Dir}(\alpha/T,\dots,\alpha/T)$ with the finite Dirichlet distribution for large $T$. For further details, see \citet{Muliere1996,Ishwaran2002}. We opt for the latter suggestion as Dirichlet weights can be generated efficiently, which leads to a simpler variant of the posterior bootstrap algorithm as shown in Algorithm \ref{Postsamp}.

\begin{algorithm}[H]
 \caption{Posterior Bootstrap Sampling}
 \label{Postsamp}
     \begin{algorithmic}
     \STATE{Define $T$ as truncation limit}
     \STATE{Observed samples are $y_{1:n}$}
       \FOR{$i=1$ {\bfseries to} $B$}
       \STATE{Draw prior pseudo-samples $\tilde{y}^{(i)}_{1:T} \overset{iid}{\sim} F_\pi$}
       \STATE{Draw $(w^{(i)}_{1:n},\tilde{w}^{(i)}_{1:T}) \sim \text{Dir}\left(1,\dots,1,\alpha / T, \dots, \alpha/T \right) $}
       \STATE{$\theta^{(i)}=  \argmin_{\theta} \left\{ \sum_{j=1}^{n} w^{(i)}_j  l(y_j,\theta)\right.$}
       \STATE{$\hspace{25mm} \left. +\sum_{k=1}^T \tilde{w}^{(i)}_k  l(\tilde{y}^{(i)}_k\negthickspace,\theta)\right\} $}
       \ENDFOR
    \end{algorithmic}
\end{algorithm}
\vskip -0.15in
For $\alpha = 0$, we simply draw $w^{(i)}_{1:n} \sim \text{Dir} \left(1,\dots,1\right)$, which is no longer an approximation and is equivalent to the Bayesian bootstrap. For $\alpha>0$, the sampling scheme is asymptotically exact for $T \to \infty$, but this is computationally infeasible. We could fix $T$ to a moderate value, or select it adaptively via adaptive NPL, where we use the stick-breaking construction until the remaining probability is less than $\epsilon$.

\subsection{Tackling Multimodal Posteriors with Initialization}

Multimodal posteriors can arise in Bayesian inference if the likelihood function is  non-log-concave like in GMMs \cite{Jin,Stephens1999}, or if the prior is non-log-concave which can arise when selecting sparse priors \cite{Seeger,Park2008,Lee2010}. Unlike the method by \citet{Lyddon} with the MDP, our NPL posterior with the DP is now decoupled from the Bayesian posterior. There is thus no reliance on an accurate  representation of the Bayesian posterior with potential multimodality, which MCMC and VB can often struggle to capture. 
If our loss function in (\ref{eq1}) is non-convex (e.g. $-\log f_\theta(y)$ of a GMM), our NPL posterior may also be multimodal. This now presents an optimization issue: solving ($\ref{eq1}$) requires non-convex optimization. In general, optimizing non-convex objectives is difficult (see \citet{nonconvex}), but under smoothness assumption of the loss, we can apply convex optimization methods to find local minima.

\subsubsection{Random Restart for Multiple Modes}
 \vspace{-0.05in}
Random restart (see \citet{Boender}) can be utilized with convex optimization methods to generate a list of potential global minima then selecting the one with the lowest objective. This involves $R$ random initializations of $\theta^{\text{init}} \sim \pi_0$ for each local optimization, and it was shown by \citet{Hu1994} that the uniform measure for $\pi_0$  has good properties for convergence. If the number of modes is finite, then the global minimum will be achieved asymptotically in the limit of the $R \to \infty$. The probability  of obtaining the correct global minimum for finite $R$ is related to the size of its basin of attraction. Random restart NPL (RR-NPL) is shown in Algorithm \ref{RRNPLsamp}.
 \vspace{-0.05in}
\begin{algorithm}[H]
 \caption{RR-NPL Posterior Sampling}
\label{RRNPLsamp}
     \begin{algorithmic}
       \FOR{$i=1$ {\bfseries to} $B$}
       \STATE{Draw $F^{(i)} \sim \text{DP} (\alpha+n, G_n) $}
       \FOR{$r=1$ {\bfseries to} $R$}
       \STATE{Draw $\theta^{\text{init}}_r \sim \pi_0$}
       \STATE{$\theta^{(i)}_r =  \localargmin_{\theta} \left( \int l(y,\theta) dF^{(i)}(y), \theta^{\text{init}}_r\right)$}
       \ENDFOR
       \STATE{$\theta^{(i)} = \argmin_r \int l(y,\theta_r^{(i)}) dF^{(i)}(y)  $}
       \ENDFOR
    \end{algorithmic}
\end{algorithm}
\vskip -0.2in
This is particularly suited to NPL with non-convex loss functions for the following reasons. Firstly, random restart can utilize efficient convex optimization techniques such as quasi-Newton methods, and the restarts can be easily implemented in parallel which is coherent with our  parallelizable sampling scheme. Secondly, we can compromise between accuracy and computational cost by selecting $R$, as computational cost scales linearly with $R$ (though we can parallelize). The repercussions of an insufficiently large $R$ are not severe: our NPL posterior will incorrectly allocate more density to local modes/saddles but all modes will likely still be present for a sufficiently large $B$. This is demonstrated in Section \ref{more_GMM} of the Supplementary Material.  Finally, the uniform initialization can sample from nonidentifiable posteriors with symmetric modes as their basins of attraction are selected with equal probability. 

Practically, uniform initialization may not be possible if the support of the parameter is infinite, e.g. the variance $\sigma^2$. In this case, we can  pick another $\pi_0$ (e.g. Gamma for a positive parameter), or sample uniformly from a truncated support. For adaptively setting $R$, we can utilize stopping rules as discussed in Section \ref{stopping_r} of the Supplementary Material.

\subsubsection{Fixed Initialization for Local Modes}
 \vspace{-0.05in}
We may be interested in targeting local modes of the posterior when we value interpretability of posterior quantities over exact posterior representation. For example in $K$-component mixture models, there will be $K!$ symmetrical modes (or sets of modes), and label-switching occurs if the sampler travels between these \cite{Jasra2005}  which impedes useful inference in terms of clustering. 

We can target one NPL posterior mode through a fixed initialization scheme by taking advantage of the fact that local optimization methods like expectation-maximization (EM) or gradient ascent are hill-climbers. We initialize each  maximization step with the same  $\theta^{\text{init}}$, causing the sampler to stay within the basin of attraction of the local posterior mode with high probability. We can utilize VB's mode-selection to select $\theta^{\text{init}}$, assuming the Bayesian and NPL posterior modes are close. Mean-field VB also tends to underestimate posterior variance \cite{Blei2017}, so we are able to obtain accurate local uncertainty quantification of the mode through this scheme. Fixed initialization NPL (FI-NPL) is shown in Algorithm \ref{MLENPLsamp}.
\vspace{-0.05in}
\begin{algorithm}[H]
 \caption{FI-NPL Posterior Sampling}
\label{MLENPLsamp}
     \begin{algorithmic}
     \STATE{Select $\theta^{\text{init}}$ from mode of interest}
       \FOR{$i=1$ {\bfseries to} $B$}
       \STATE{Draw $F^{(i)} \sim \text{DP} (\alpha+n, G_n) $}
       \STATE{$\theta^{(i)} =  \localargmin_{\theta} \left( \int l(y,\theta) dF^{(i)}(y), \theta^{\text{init}}\right)$}
   	 \ENDFOR
    \end{algorithmic}
\end{algorithm}
\vspace{-5mm}

\subsection{Loss-NPL} \label{loss}

As we cannot define priors on $\theta_0$ directly, we can instead penalize undesirable properties in the loss
\begin{equation} \label{eq6}
l(y,\theta) = -\log f_\theta (y) + \gamma g(\theta).
\end{equation}
For example, $g(\theta) = |\theta|$ obtains the Bayesian NPL-Lasso, or we can set $g(\theta) = -\log \pi(\theta)$ if we have some prior preference. We recommend  $\gamma = \frac{1}{n}$ if we desire roughly the same prior regularization as in Bayesian inference, where $n$ is the size of the training set. The reasoning is outlined in Section \ref{loss_sup} of the Supplementary Material. We could also tune $\gamma$ through desired predictive performance or properties of $\theta$. Note that we are no longer encoding prior beliefs, and are instead expressing an alternative parameter of interest that minimizes the expectation of (\ref{eq6}).

\subsection{Related Work}

We build on the work of \citet{Lyddon} which specifies an MDP prior on $F_0$, and recovers conventional Bayesian inference in the limit of $\alpha \to \infty$.  Although the foundations of nonparametric learning are unchanged, our NPL posterior is decoupled from the Bayesian model, offering flexibility in prior measure selection, computational scalability and full multimodal exploration.

NPL unsurprisingly overlaps with other nonparametric approaches to inference. We recover the Bayesian bootstrap \cite{Rubin1981} if we set $\alpha = 0$, and further setting $l(y,\theta) = -\log f_\theta (y)$ gives the weighted likelihood bootstrap \cite{Newton1994}, as discussed in \citet{Lyddon2019}. Setting the loss to ($\ref{eq6}$) and $\alpha  = 0$ also returns the fixed prior weighted Bayesian bootstrap \cite{Newton2018}. However, these methods were posited as approximations to the true Bayesian posterior, and the Bayesian bootstrap/weighted likelihood bootstrap are unable to incorporate prior information. The NPL posterior on the other hand is exact and distinct to the conventional Bayesian posterior with theoretical advantages, and we are able to incorporate prior information either through $F_\pi$ or $l(y,\theta)$.

Treating parameters as functionals of the sampling distribution is akin to empirical likelihood methods \cite{Owen1988}, in which parameters are defined through estimating equations of the form $\int m(y,\theta) dF_0(y) = 0$. The definition of a parameter of interest through the loss $l(y,\theta)$ is also present in general Bayesian updating introduced by \citet{Bissiri2016}, where a coherent posterior over a parameter of interest is obtained without the need to specify a joint generative model. Their target parameter is equivalent to (\ref{eq1}), and their methodology is built on a notion of coherency. 

\section{Examples}\label{examples}

We now demonstrate our method on some examples; the code is available online \footnote{\url{https://github.com/edfong/npl}}. We compare NPL to conventional Bayesian inference with the No-U-Turn Sampler (NUTS) by  \citet{Homan2014}, and Automatic Differentiation Variational Inference (ADVI) by \citet{Kucukelbir2017} in Stan \cite{Carpenter_stan:a}. We select these as baselines as they are off-the-shelf algorithms that do not require tuning. Similarly, NPL only requires a weighted likelihood optimization procedure. All NPL examples are run on 4 Azure F72s\_v2 (72 vCPUs) virtual machines, implemented in Python.  The NUTS and ADVI examples cannot be implemented in an embarrassingly parallel manner, so they are run on a single  Azure F72s\_v2. We avoid running multiple MCMC chains in parallel as the models are multimodal which may impede mixing, and combining unmixed chains is unprincipled.  For tabulated results, each run was repeated 30 times with different seeds, and we report the mean with 1 standard error.  We emphasize again that our NPL posterior is distinct to the conventional Bayesian posterior,  so we are comparing the two inference schemes and their associated sampling methods. We include additional empirical comparisons to importance sampling and NPL with an MDP prior in Sections \ref{Imp}, \ref{MDP}  of the Supplementary Material. 

\subsection{Gaussian Mixture Model}\label{GMMsec}

We demonstrate the ability of RR-NPL  to accurately sample from a multimodal posterior in a $K$-component, $d$-dimensional diagonal GMM toy problem, which NUTS and ADVI fail to do. It should be noted that in addition to the $K!$ symmetrical modes present from label-switching, further multimodality is present due to the non-log-concavity of the likelihood. We further show how FI-NPL can be used in a clustering example with real data to provide accurate local uncertainty quantification which ADVI is unable to do. Our conventional Bayesian model for $i \in \{1, \dots,n \}$, $j \in \{1, \dots,d \}$ and $k \in \{1, \dots,K \}$ is
 \vspace{-0.02in}
\begin{equation} \label{GMM}
\begin{aligned}
\mathbf{y}_i | \boldsymbol{\pi},\boldsymbol{\mu},\boldsymbol{\sigma} &\sim \sum_{k=1}^K \pi_k \mathcal{N}\left(\boldsymbol{\mu}_k, \text{diag}(\boldsymbol{\sigma}_k^2 )\right), \\
\boldsymbol{\pi} | a_0 &\sim \text{Dir}(a_0,\dots,a_0),\\
\mu_{kj} &\sim \mathcal{N}(0,1), \\
\sigma_{kj} &\sim \text{logNormal}(0,1).
\end{aligned}
\end{equation}
The posterior is multimodal, and we use ADVI and NUTS for inference. For NPL, we are interested in model fitting, so our loss function is simply the negative log-likelihood
 \vspace{-0.02in}
\begin{equation}
l(\mathbf{y},\boldsymbol{\pi},\boldsymbol{\mu},\boldsymbol{\sigma}) = -\log \sum_{k=1}^K \pi_k \mathcal{N}\left(\mathbf{y}; \boldsymbol{\mu}_k, \text{diag}(\boldsymbol{\sigma}_k^2 )\right).
\end{equation}
In the case of small $n$, we may want to include a regularization term in the loss  to avoid singularities of the likelihood. We select the DP prior separately for each example. 

\begin{figure*}[!ht]
\begin{center}
\centerline{\includegraphics[width=1.9\columnwidth]{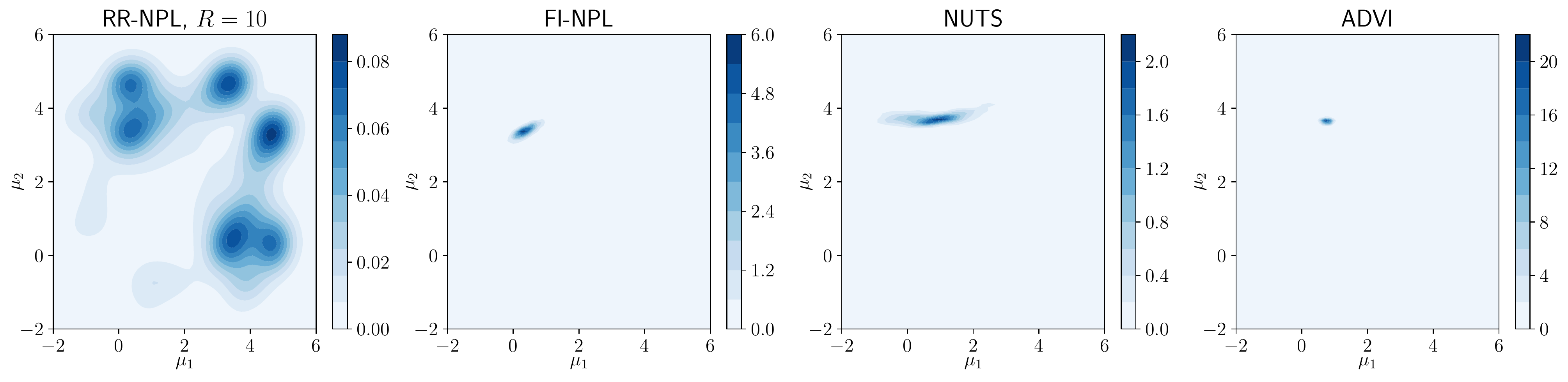}}
\vskip -0.1in
\caption{Posterior KDE of $(\mu_1,\mu_2)$ in $K=3$ toy GMM problem}
\label{mu}
\end{center}
\vskip -0.3in
\end{figure*}

\subsubsection{Toy Example: Implementation and Results}
 \vspace{-0.05in}
We analyze toy data from a GMM with $K=3$, $d = 1$ and the following parameters:
\begin{equation*}
\begin{aligned}
\boldsymbol{\pi}_0 = \{0.1,0.3,0.6\}, \hspace{2mm} \boldsymbol{\mu}_0 = \{0,2,4\}, \hspace{2mm} \boldsymbol{\sigma}_0^2 = \{1,1,1\}.
\end{aligned}
\end{equation*}
We generate $n_{\text{train}} = 1000$ for model fitting and another $n_{\text{test}}=250$ held-out for model evaluation with different seeds for each of the 30 runs. For the Bayesian model we set $a_0=1$, and for NPL we set $\alpha = 0$ as $n \gg p$.  We optimize each bootstrap maximization with a weighted EM algorithm (derived in Section \ref{weighted_EM} of the Supplementary Material), and implement this in a modified \texttt{GaussianMixture} class from  \texttt{sklearn.mixture} \cite{scikit-learn}. For RR-NPL, we initialize $\boldsymbol{\pi} \sim \text{Dir}(1,\dots,1)$, $\mu_{kj} \sim \text{unif}(-2,6)$ and $\sigma^2_{kj}\sim \text{IG}(1,1)$ for each restart. For FI-NPL we initialize with one of the posterior modes from RR-NPL. We produce $2000$ posterior samples for each method. We evaluate the predictive performance of each method on held-out test data with the mean log pointwise predictive density (LPPD) as suggested by \citet{gelman2013bayesian}, which is described in Section \ref{pred_perf} of the Supplementary Material. A larger value is equivalent to a better fit to the test data.

Figure \ref{mu} shows the posterior KDEs of ($\mu_1,\mu_2$) for 1 run of each method. RR-NPL clearly recovers the multi-modality of the NPL posterior, including the symmetry about $\mu_1 = \mu_2 $ due to the nonidentifiability of the GMM posterior.  NUTS and ADVI remain trapped in one local mode of the Bayesian posterior as expected. Even if we carried out random initialization of NUTS/ADVI over multiple runs, each run would only pick out one mode, and there is no general method to combine the posteriors.  ADVI also clearly underestimates the marginal posterior uncertainty.  FI-NPL remains in a single mode, showing that we can fix label-switching through this initialization. However, the FI-NPL mode is not identical to a truncated version of the RR-NPL mode, as posterior mass is not reallocated symmetrically from the other modes. We see in Tables \ref{GMMLP}, \ref{GMMruntime} that RR-NPL has similar mean LPPD on toy test data compared to NUTS, and is twice as fast as NUTS.

\subsubsection{MNIST: Implementation and Results}
\begin{figure}[H]
\begin{center}
\centerline{\includegraphics[width=\columnwidth]{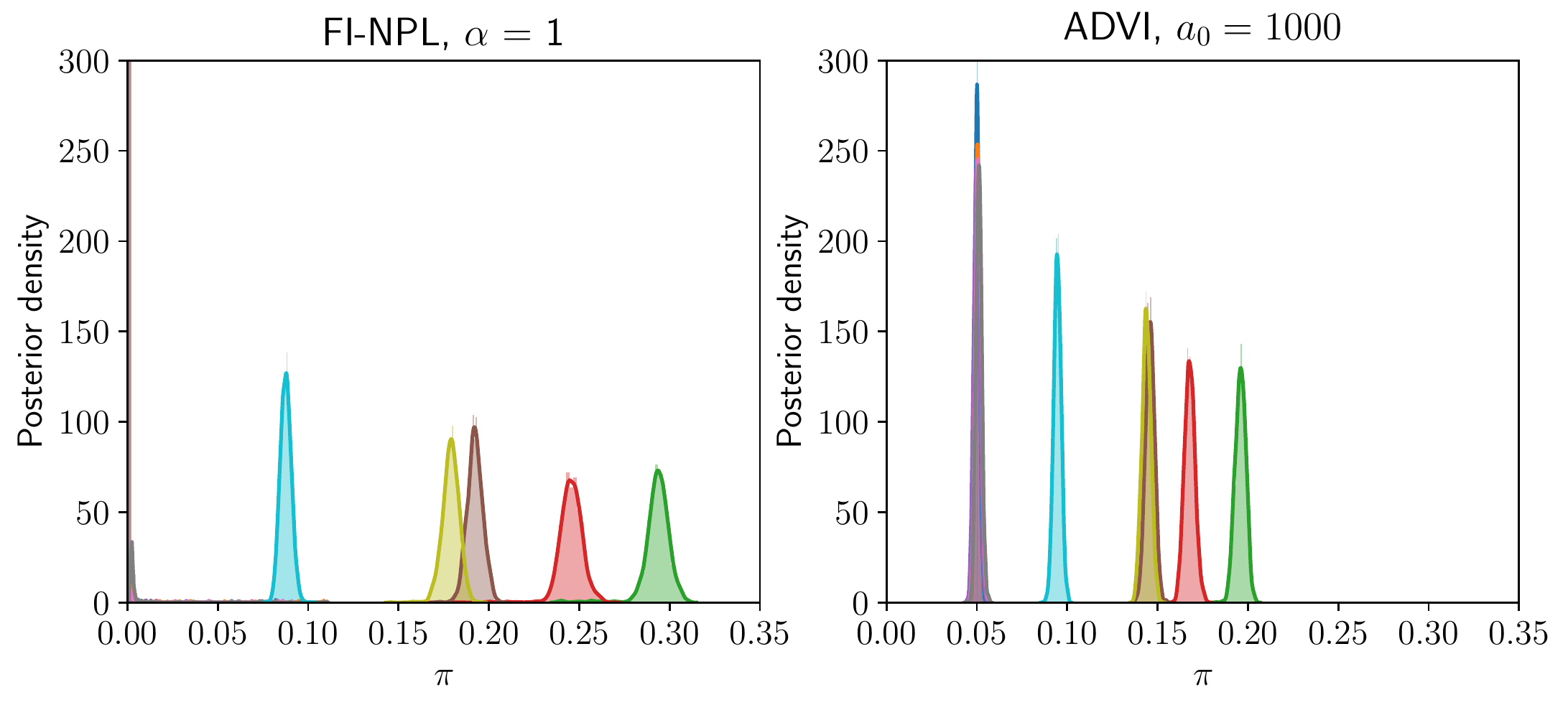}}
\vskip -0.1in
\caption{Posterior marginal KDEs of $\boldsymbol{\pi}$ for K=10 GMM on MNIST; 5 of the components have been set to 0 for FI-NPL, and likewise to 0.05 for ADVI}
\label{VBinit}
\end{center}
\vskip -0.35in
\end{figure}

 \vspace{-0.05in}
We now demonstrate FI-NPL on clustering handwritten digits from MNIST \cite{mnist}, which consists of $28 \times 28$ pixel images. In this example $n_{\text{train}} = 10000, n_{\text{test}} = 2500$ and $d=784$. We normalize all pixel values such that they lie in the interval $[0,1]$, and set $K=10$. We believe a priori that many pixels are close to 0, so for ease we elicit a tight normal centering measure for the DP
\begin{equation}
f_\pi(\mathbf{y}) =  \prod_{j=1}^d \mathcal{N}(y_j; 0, 0.1^2).
\end{equation}
NUTS is prone to the label-switching problem and is too computationally intensive as ADVI already requires 5 hours, so we only compare FI-NPL to ADVI. We set $a_0 = 1000$ for ADVI, and $\alpha=1$ for FI-NPL with $T=500$. We carry out a single run of ADVI to select a local mode, and set $\theta^{\text{init}}$ of FI-NPL to the ADVI-selected mode. We then carry out 30 repeats of FI-NPL with this initialization, and compare to the original ADVI run.  We see in Figure \ref{VBinit} that we obtain larger posterior variances in FI-NPL, as ADVI likely underestimates the posterior variances due to the mean-field approximation. Notice the modes are not exactly aligned as the NPL and Bayesian posterior are distinct, and furthermore ADVI is approximate. We conjecture that ADVI does not set components exactly to 0 due to the strong Dirichlet prior. We see in Tables \ref{GMMLP}, \ref{GMMruntime} that FI-NPL is predictively better and runs around 300 times faster than ADVI.

\begin{table}[H]
\vskip -0.1in
\caption{Mean LPPD on held-out test data for GMM}
\label{GMMLP}
\vskip -0.1in
\begin{center}
\begin{tiny}
\begin{sc}
 \resizebox{\columnwidth}{!}{
\begin{tabular}{c||c|c|c|c}
  &RR-NPL& FI-NPL &NUTS &ADVI   \\
\hline \hline
Toy &\textbf{-1.909}$\pm$ 0.040& -1.911 $\pm$ 0.040&  -\textbf{1.908 }$\pm$0.039 &-1.912 $\pm$ 0.041\\
 MNIST &/&\textbf{2463.4} $\pm$24.1&/& 1188.2\\
\end{tabular}
}
\end{sc}
\end{tiny}
\end{center}
\vskip -0.4in
\end{table}

\begin{table}[H]
\caption{Run-time for 2000 samples for GMM}
\label{GMMruntime}
\vskip 0.05in
\begin{center}
\begin{tiny}
\begin{sc}
\begin{tabular}{c||c|c|c|c}
  & RR-NPL& FI-NPL &NUTS&ADVI   \\
\hline \hline
 Toy&37.2s $\pm$ 4.5s &5.5$\pm$ 2.2s &1m20s  $\pm$ 16s &  0.8s $\pm$ 0.1s      \\
 MNIST &/&57.9s $\pm$ 1.0s&/& 5h6m\\
\end{tabular}
\end{sc}
\end{tiny}
\end{center}
\vskip -0.3in
\end{table}

\subsection{Logistic Regression with Automatic Relevance Determination Priors} \label{ARD}

We now demonstrate the predictive performance and computational scalability of loss-NPL in a Bayesian sparse logistic regression example on real datasets. To induce sparsity, we place automatic relevance determination (ARD) priors \cite{MacKay94} on the coefficients with Gamma hyperpriors \cite{Gelman2008}. The conventional Bayesian model for $i \in \{1, \dots,n \}$ and $j \in \{1, \dots,d \}$ is
\begin{equation} \label{logreg}
\begin{aligned}
y_i | \mathbf{x}_i ,\boldsymbol{\beta},\beta_0 &\sim \text{Bernoulli}(\eta_i), \\ \eta_i &= \sigma({ \boldsymbol{\beta}}^T \mathbf{{x}}_i + \beta_0), \\ \beta_j | \lambda_j &\sim \mathcal{N}\left(0, \frac{1}{\lambda_j}\right),\\ \lambda_j|a,b &\sim \text{Gamma}\left(a,b \right). \end{aligned}
\end{equation}
Marginally, the prior is the non-standardized t-distribution with (degrees of freedom, location, squared scale)
\begin{equation}
\beta_j \sim \text{Student-t}\left(2a, 0, \frac{b}{a} \right).
\end{equation}
This posterior is intractable and potentially multimodal due to the non-log-concavity of the prior, and we carry out conventional Bayesian inference via NUTS and ADVI.  When applying loss-NPL to regression, we assume $y,x \overset{iid}{\sim} F_0$, and place a DP prior on the joint distribution $F_0(y,x)$. We target the parameter which satisfies (\ref{eq1}) with loss
\begin{equation}
\begin{aligned}
l(y,\mathbf{x},\boldsymbol{\beta},\beta_0) = -\left(y \log \eta + (1-y) \log(1-\eta) \right) \\ + \gamma\left(\frac{2a+1}{2}\right) \sum_{j=1}^d  \log \left(1 + \frac{\beta_j^2}{2b} \right)
\end{aligned}
\end{equation}
which is the negative sum of the log-likelihood and log-prior, with additional scaling parameter $\gamma$. Again our NPL posterior may be multimodal due to the non-convexity of the loss, and so we utilize RR-NPL. It should be noted that our target parameter is now different to conventional Bayesian inference, but our method achieves the common goal of variable selection under a Bayesian framework. For the DP prior, we elicit the centering measure
\begin{equation}
\begin{aligned}
f_\pi(y,x) &= f_\pi (y) f_\pi(x),\\
f_\pi(y) &= \text{Bernoulli}(0.5),\\
f_\pi(x) &= \frac{1}{n}\sum_{i=1}^n \delta_{x_i}(x).
\end{aligned}
\end{equation}
The prior assumes $y,x$ are independent which is equivalent to assuming $ \boldsymbol{\beta} = \mathbf{0}$ a priori. This  is appropriate as we believe many components of $ \boldsymbol{\beta}$ to be close to 0. The prior on $x$ is its empirical distribution, which is in an empirical Bayes manner where the prior is estimated from the data. 

\subsubsection{Implementation and Results \label{ARDimp}}
We analyze 3 binary classification datasets from the UCI ML repository \cite{Dua:2017}: `Adult' \cite{Kohavi96scalingup}, `Polish companies bankruptcy 3rd year', \cite{zikeba2016ensemble}, and `Arcene' \cite{Guyon2005} with details in Table \ref{UCI}. We handle categorical covariates with dummy variables, and normalize all covariates to have mean 0 and standard deviation 1. Missing real values were imputed with the mean, and data  with missing categorical values were dropped. We carry out a random stratified train-test split for each of the 30 runs, with 80-20 split for `Adult', `Polish' and 50-50 split for `Arcene' due to the smaller dataset.  For both NPL and conventional Bayesian inference, the hyperparameters were set to $a=b=1$, which was selected by tuning the sparsity of the Bayesian posterior means to a desired value. For NPL, we set $\alpha = 0$ for `Adult' and `Polish' as $n$ is sufficiently large, and  $\alpha=1$ for `Arcene' with $T=100$ as $n$ is only $100$. We set $\gamma = \frac{1}{n_{\text{train}}}$ for each dataset as explained in Section \ref{loss} for a fair comparison to the conventional Bayesian model. We initialize each optimization with  $\beta_j^0 \sim \mathcal{N}(0,1)$, and select the number of restarts to $R=1$ for expediency. Optimization was carried out using the L-BFGS-B algorithm \cite{Zhu1997} implemented in \texttt{scipy.optimize} \cite{scipy}. 

We can see in Table \ref{LPPD} that loss-NPL is predictively similar or better than NUTS and ADVI, and from Table \ref{sparsity} we see that the posterior mean is sparser for loss-NPL. Finally, we see from Table \ref{runtime} that the loss-NPL run-times for $2000$ posterior samples are much faster than for NUTS, and comparable to VB. Further  measures of predictive performance are provided in Section \ref{more_logreg} of the Supplementary Material.

\begin{table}[H]
\vskip -0.1in
\caption{UCI datasets descriptions for LogReg}
\label{UCI}
\vskip 0.05in
\begin{center}
\begin{scriptsize}
\begin{sc}
\begin{tabular}{c||c|c|c|c|c}
 Data Set & Type &$d$ & $n_{\text{train}}$&$n_{\text{test}}$ & Positive \%   \\
\hline \hline
Adult& Cat. &96&36177&9045 & 24.6 \\
Polish&Real&64&8402&2101&4.8\\
Arcene&Real&10000&100&100&44.0 
\end{tabular}
\end{sc}
\end{scriptsize}
\end{center}
\vskip -0.4in
\end{table}

\begin{table}[H]
\caption{Mean LPPD on held-out test data for LogReg}
\label{LPPD}
\vskip 0.05in
\begin{center}
\begin{scriptsize}
\begin{sc}
\begin{tabular}{c||c|c|c}
 Data Set & Loss-NPL & NUTS &ADVI   \\
\hline \hline
 Adult &\textbf{ -0.326}$\pm$0.004 & \textbf{-0.326}$\pm$0.004   & -0.327 $\pm$ 0.004\\
 Polish & \textbf{-0.229}$\pm$ 0.034 & -3.336$\pm$ 4.162& -0.247 $\pm$  0.047\\
 Arcene &\textbf{-0.449} $\pm$ 0.104&-0.464 $\pm$ 0.032&\textbf{-0.445} $\pm$0.068
\end{tabular}
\end{sc}
\end{scriptsize}
\end{center}
\vskip -0.4in
\end{table}

\begin{table}[H]
\caption{Percentage of posterior mean $|\beta_j|< \epsilon$ for LogReg}
\label{sparsity}
\vskip 0.05in
\begin{center}
\begin{scriptsize}
\begin{sc}
\begin{tabular}{c|c||c|c|c}
 Data Set &$\epsilon$& Loss-NPL & NUTS & ADVI  \\
\hline \hline
 Adult & 0.1 & \textbf{17.6} $\pm$ 2.8 & 16.1 $\pm$  2.7 &12.1$\pm$  3.1 \\
 Polish &0.1&\textbf{33.5} $\pm$ 4.7 &15.9 $\pm$3.3 & 15.8$\pm$3.5\\
 Arcene &0.01&\textbf{87.4} $\pm$ 0.7 &4.7 $\pm$ 0.3 &3.5 $\pm$ 0.3
\end{tabular}
\end{sc}
\end{scriptsize}
\end{center}
\vskip -0.4in
\end{table}

\begin{table}[H]
\caption{Run-time for 2000 samples for LogReg}
\label{runtime}
\vskip 0.05in
\begin{center}
\begin{scriptsize}
\begin{sc}
\begin{tabular}{c||c|c|c}
 Data Set & Loss-NPL & NUTS & ADVI\\
\hline \hline
 Adult & 2m24s $\pm$ 8s &2h36m $\pm$ 4m & 26.9s $\pm$ 7.3s  \\
 Polish &19.0s $\pm$ 4.0s &1h20m $\pm$ 21m &3.3s $\pm$ 0.8s  \\
 Arcene &53.5s $\pm$ 1.1s &4h31m $\pm$  53m&54.2s $\pm$3.3s
\end{tabular}
\end{sc}
\end{scriptsize}
\end{center}
\vskip -0.2in
\end{table} 

\subsection{Bayesian Sparsity-path-analysis }
We now utilize loss-NPL to carry out Bayesian sparsity-path-analysis for logistic regression, which allows us to visualize how the responsibility of each covariate changes with the sparsity penalty as discussed by \citet{Lee2012BayesianSO}.  We use the same ARD prior as Section \ref{ARD} with the same initialization scheme, set $\gamma = \frac{1}{n}$, and elicit a noninformative DP prior with $\alpha = 0$. We found empirically that the results for larger values of $R$ are similar and so the approximation with $R=1$ is sufficient.  We fix $a$ and vary the value of $b$ to favour solutions of different sparsity. This varies the squared scale $c=b/a$ of the Student-t prior with fixed degrees of freedom, where a smaller $c$ corresponds to a heavier sparsity penalty and thus more components are set to 0.

\subsubsection{Implementation and Results}

\begin{figure}[ht]
\vskip -0.1in
\begin{center}
\centerline{\includegraphics[width=0.5\columnwidth]{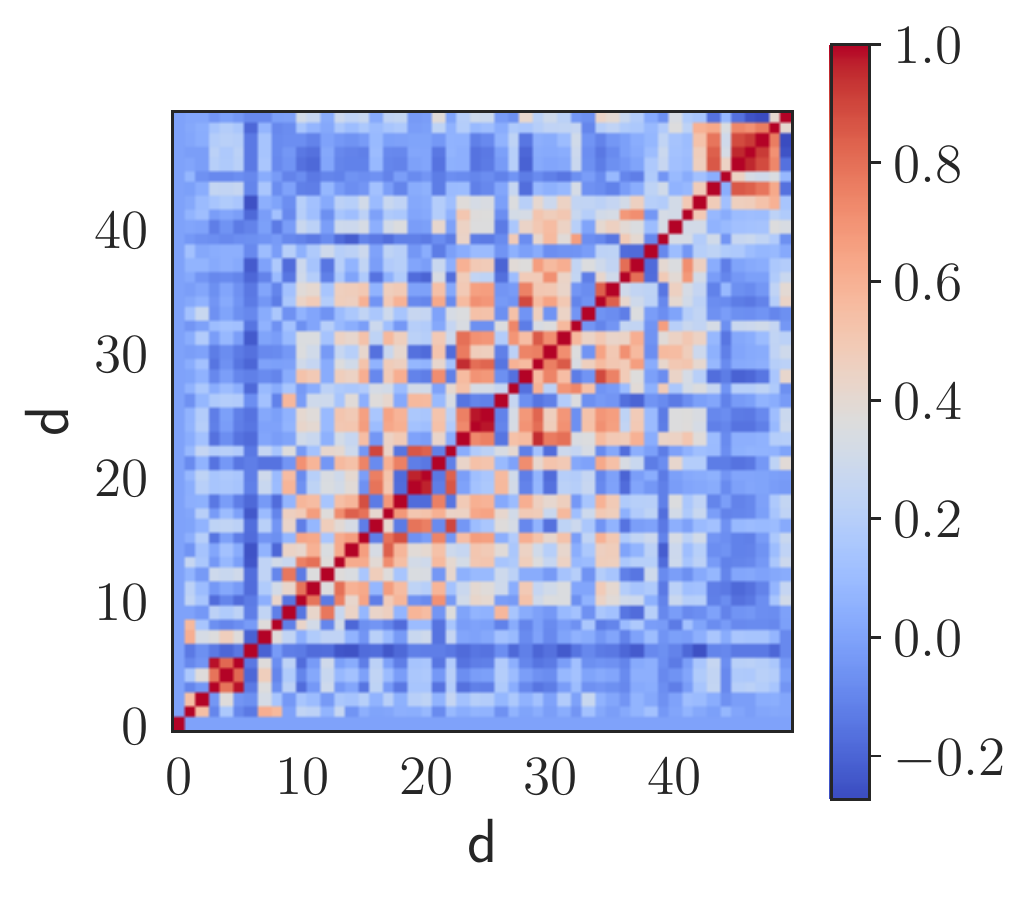}}
\vskip -0.1in
\caption{Correlations of covariates $\boldsymbol{x}$ from genetic dataset}
\label{correlations}
\end{center}
\vskip -0.3in
\end{figure}

We analyze the genotype/pseudo-phenotype dataset with $n=500$ as described by \citet{Lee2012BayesianSO},  containing patient covariates  $\mathbf{x}_i$ which exhibit strong block-like correlations as shown in Figure \ref{correlations}. We normalize the covariates to have mean 0 and standard deviation 1. The pseudo-phenotype data is generated by $y_i \sim \text{Bernoulli}(\sigma({ \boldsymbol{\beta}}^T \mathbf{{x}}_i) )$, where $\boldsymbol{\beta}$  has 5 randomly selected non-zero components out of $d=50$, with the rest set to 0. Each non-zero component is sampled from $\mathcal{N}(0,0.2)$, and the exact values of  $\boldsymbol{\beta}$ are provided in Section \ref{beta_vals} of the Supplementary Material. We set $a=1$ and vary $b_t = 0.98^{t-1}$ for $t = \{1, \dots , 450 \}$, and generate 4000 posterior samples for each setting.

\begin{figure}[ht]
\vskip 0.05in
\begin{center}
\centerline{\includegraphics[width=\columnwidth]{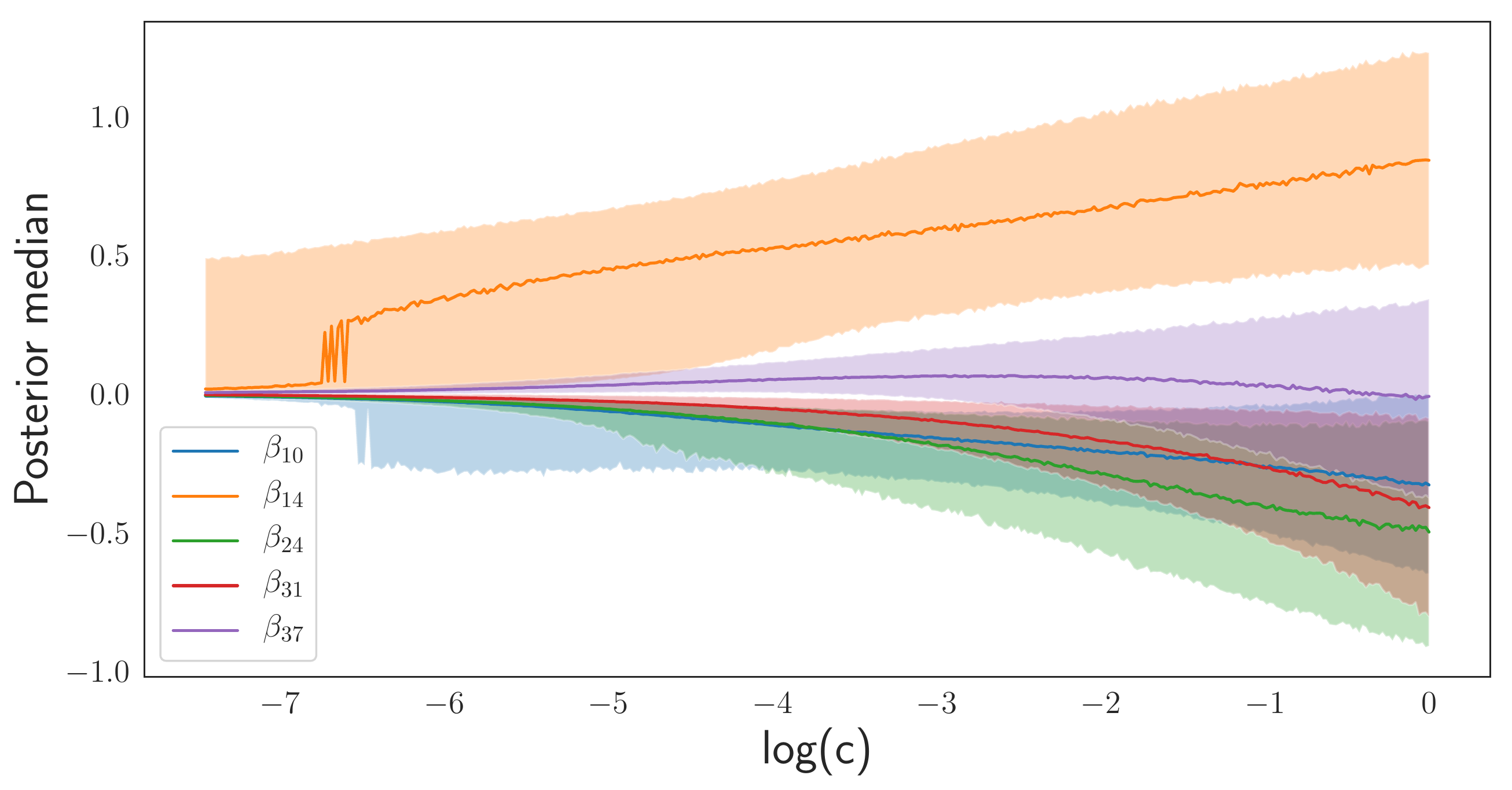}}
\vskip -0.05in
\caption{Lasso-type plot for posterior medians of non-zero $\boldsymbol{\beta}$ with 80\% credible intervals against $\log(c)$  from genetic dataset }
\label{sparsemedian}
\end{center}
\vskip -0.2in
\end{figure}

\begin{figure}[!ht]
\begin{center}
\centerline{\includegraphics[width=\columnwidth]{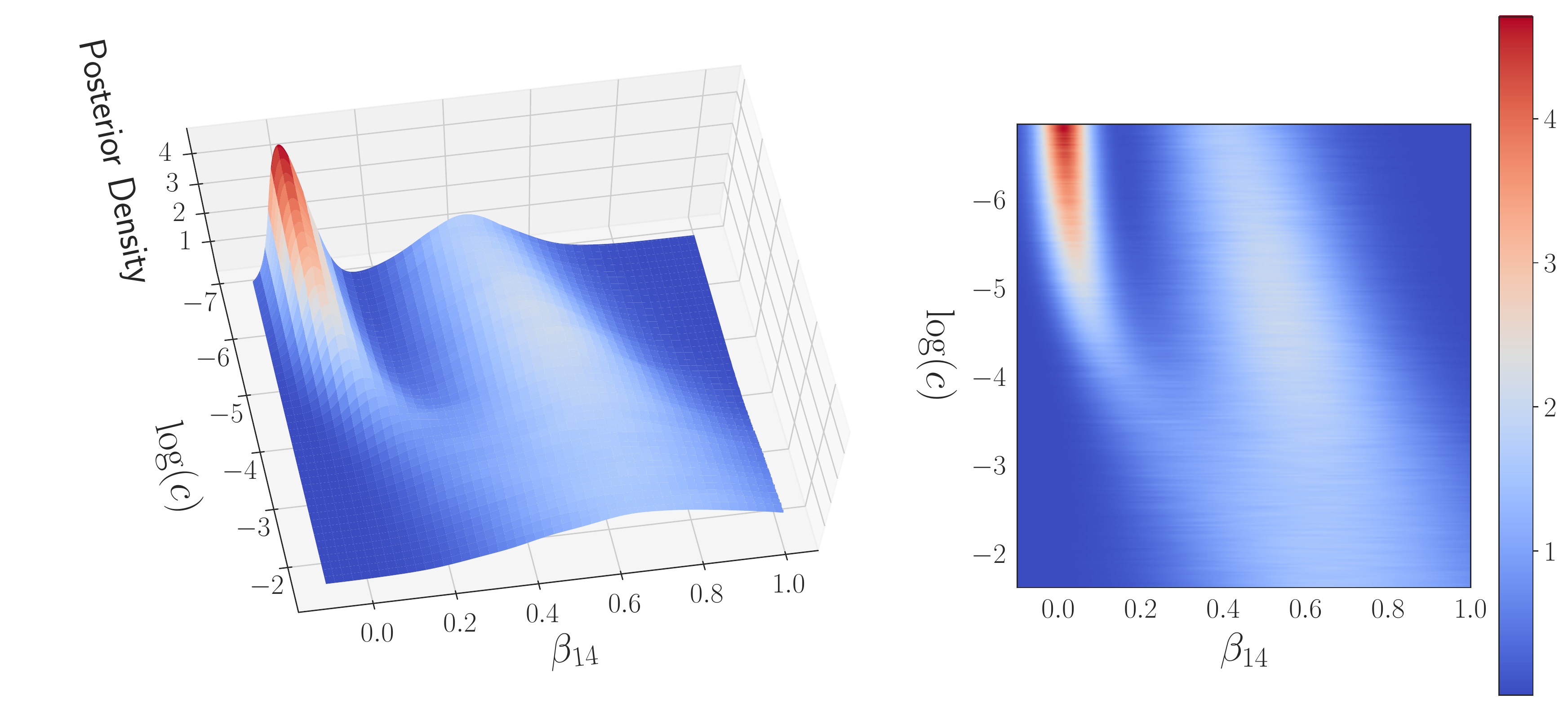}}
\vskip -0.1in
\caption{Posterior marginal KDE of $\beta_{14}$ against $\log(c)$ from genetic dataset}
\label{surf}
\end{center}
\vskip -0.3in
\end{figure}

The posterior medians of the non-zero components of $\boldsymbol{\beta}$ with $80\%$ central credible interval are shown in Figure \ref{sparsemedian} for a range of $\log c$ values. Both the posterior median and central credible intervals are estimated through the appropriate order statistics of the posterior samples \cite{gelman2013bayesian}. We can see that $\beta_{10}$, $\beta_{14}$ and $\beta_{24}$ have early predictive power as their credible intervals remain large despite a significant sparsity penalty (small $\log(c)$), whilst the other two coefficients  $\beta_{31},\beta_{37}$ are masked. A plot of the absolute medians for all components is included in Section \ref{more_gen} of the Supplementary Material.  For $\beta_{10}$ and $\beta_{14}$, the median is close to 0 but the credible interval is large which is due to the multimodality of the marginal posterior. This multimodality is also responsible for the jitter in the median around $\log(c)= -6.5$ for $\beta_{14}$ in Figure \ref{sparsemedian} ; the true median likely lies between the two separated modes but the finite posterior sample size  causes the sample median to jump between the two. A posterior marginal KDE plot of $\beta_{14}$ changing with $\log c$ is shown in Figure \ref{surf}, allowing us to visualize how the importance of the covariate changes with the sparsity penalty. We observe the bimodality in the marginal posterior for $\log(c) < -4$ as expected from the  above discussion.

Loss-NPL required 5 minutes 24 seconds to generate all $450 \times 4000$ posterior samples. The computational speed of NPL enables fast Bayesian analysis of large datasets with different hyperparameter settings, allowing for Bayesian variable selection analysis. 


\section{Discussion}

We have introduced a variant of Bayesian nonparametric learning (NPL) with a Dirichlet process (DP) prior on the sampling distribution $F_0$, which leads to highly scalable exact inference under model misspecification, detached from the conventional Bayesian posterior. 
This method admits a sampling scheme for multimodal posteriors that allows for full mode exploration, which involves a non-convex optimization that we solve through random restart. We demonstrated that NPL can perform predictively better than conventional Bayesian inference, while providing exact uncertainty quantification.

For future work, the small sample performance of NPL could be further explored and compared to conventional Bayesian inference; we currently recommend NPL for moderate to large values of $n$.  The scaling of the  number of repeats $R$ with increasing dimension for full mode exploration  would also be a future avenue of research.

\section*{Acknowledgements}
We would like to thank our anonymous reviewers for their
careful analyses of our paper and their valuable input.  We also thank Luke Kelly for useful discussions, Ho Chung Leon Law for helpful feedback and Anthony Lee for providing the genotype dataset. EF is funded by The Alan Turing Institute Doctoral Studentship, under the EPSRC grant EP/N510129/1. SL is funded by the EPSRC OxWaSP CDT, through EP/L016710/1. CH is supported by  The Alan Turing Institute, the HDR-UK, the Li Ka Shing Foundation, and the MRC.

\bibliography{icml}
\bibliographystyle{icml2019}
%

\onecolumn
\icmltitle{Supplementary Material: Scalable Nonparametric Sampling from Multimodal Posteriors with the Posterior Bootstrap}




\appendix
\numberwithin{equation}{section}

\setcounter{equation}{0}

\numberwithin{figure}{section}
\numberwithin{table}{section}
\setcounter{figure}{0}
\section{Eliciting the Prior Dirichlet Process}
\subsection{Intuition of the Prior $F_\pi$}\label{prior_intuit}
The parameter of interest when model fitting \cite{Walker2013} is

\begin{equation} \label{eq1_sup}
\begin{aligned}
\theta_0(F_0) &= \argmax_{\theta} \int \log f_\theta(y) dF_0(y)\\
&= \argmin_{\theta} \text{KL}(f_0||f_\theta).
\end{aligned}
\end{equation}

The prior on $F_0$ is
\begin{equation}\label{eq2_sup}
[F | \alpha,F_\pi ]\sim \text{DP}\left(\alpha, F_\pi \right).
\end{equation}

The effects of the implicit prior on $\theta_0$ due to $F_\pi$ when model-fitting can be seen in the limit of $\alpha \to \infty$ under regularity conditions:
\begin{equation}\label{eq3_sup}
\begin{aligned}
\theta_0 (F) \overset{p}{\rightarrow} \argmin_\theta \text{KL} (f_\pi || f_\theta).
\end{aligned}
\end{equation}

In the limit, the prior collapses on one of the points that minimizes the KL divergence between the prior centering density and the model. Intuitively, the prior regularizes $\theta_0$ towards (\ref{eq3_sup}), and $\alpha$ acts as weighting between $F_\pi$ and $F_n = \frac{1}{n}\sum_{i=1}\delta_{y_i}$. It is thus a measure of belief that $F_
\pi$ is the true sampling distribution.

\subsection{Selecting $\alpha$ through the Mean Functional}\label{mean_func}
We can tune $\alpha$ through the a priori variance of the mean functional
\begin{equation}\label{eq4_sup}
\begin{aligned}
\theta_{\mu}(F) &= \argmin_\theta \int (y-\theta)^2 dF(y)\\
&=\int y dF(y).
\end{aligned}
\end{equation}

If $F \sim \text{DP}(\alpha,F_\pi)$, then the a prior variance of (\ref{eq4_sup}) follows from the properties of the Dirichlet process (DP):
\begin{equation}
\text{Var}\left[\theta_{\mu}(F) \right] = \frac{\text{Var}_{F_\pi}\left[y \right]}{1+\alpha}
\end{equation}

and so we can elicit $\alpha$ from a priori knowledge of $\text{Var}\left[\theta_{\mu}(F) \right]$.
\section{Stopping Rules for Adaptively Selecting $R$}\label{stopping_r}

Although not explored in our paper, we can utilize heuristic stopping rules for adaptively selecting $R$ for full mode exploration when sampling from the NPL posterior. A simple example is to stop the repeats if there have been no improvements in the optimized function value for the last $m$ repeats, where $m$ is the parameter of the stopping rule. More complex methods involve estimating the missing probability mass due to local minima not being observed, and thresholding based on that. See \citet{ Betro1987, Dick} for a comparison of some methods. Although there is no clear answer for selecting $R$,  we can also parallelize over restarts to alleviate the computation burden.

\section{Stochastic Subsampling}\label{stochsamp}
For very large $n$, we can utilize stochastic gradient methods by subsampling to optimize the weighted loss. The full weighted loss and gradient are defined as
\begin{equation}
\begin{aligned}
\mathcal{L}(\theta) &= \sum_{i=1}^n w_i l(y_i,\theta),\\
\nabla_\theta {L}(\theta) &= \sum_{i=1}^n w_i \nabla_\theta l(y_i,\theta).
\end{aligned}
\end{equation}
If we subsample a mini-batch $\tilde{y}_{1:m} \overset{iid}{\sim} \sum_{i=1}^n w_i \delta_{y_i}$, we can then calculate the mini-batch gradient
\begin{equation}
\nabla_\theta {L}^m(\theta) = \frac{1}{m}\sum_{i=1}^ m  \nabla_\theta l(\tilde{y}_i,\theta).
\end{equation}
The mini-batch gradient is unbiased:
\begin{equation}
\begin{aligned}
\EX\left[\nabla_\theta {L}^m(\theta)\right] &= 
\EX\left[ \nabla_\theta l(\tilde{y},\theta)\right]
= \sum_{i=1}^n w_i \nabla_\theta l(y_i,\theta).
\end{aligned}
\end{equation}
Setting $m=1$ allows use to use stochastic gradient descent (SGD) and its variants which improves scalability. Furthermore, extensions to SGD such as ADAGRAD \cite{Duchi2011} and ADAM \cite{Kingma} help with escaping saddle points, which can potentially reduce the number of $R$ required for RR-NPL to obtain full mode exploration.

\section{Selecting $\gamma$ in Loss-NPL} \label{loss_sup}

For loss-NPL we can set the loss function to
\begin{equation}
l(y,\theta) = -\log f_\theta (y) - \gamma \log \pi(\theta).
\end{equation}
In this case, we recommend the scaling parameter to be $\gamma = \frac{1}{n}$ if we want roughly the same prior regularization of $\pi(\theta)$ as in traditional Bayesian inference. This can be seen when we look at the expected of $\int l(y,\theta) dF$ for $\alpha = 0$ (i.e. $F\sim \text{DP}\left(n, \frac{1}{n}\sum_{i=1}^n \delta_{y_i}\right)$):
\begin{equation}
\begin{aligned}
\EX\left[\int l(y,\theta) dF \right] = &-\frac{1}{n}\ \sum_{i=1}^n  \log f_\theta(y_i)- \gamma \log \pi(\theta).
\end{aligned}
\end{equation}
We obtain the same weighting as in Bayesian inference between the log-likelihood and log-prior for $\gamma = \frac{1}{n}$.
\section{Examples}

\subsection{Toy Example: Normal Location Model} \label{ex1_sup}
We now empirically demonstrate the small sample performance of NPL and  the role of the prior concentration $\alpha$ in a toy normal location model problem. Suppose the model of interest is $f_\theta (y) = \mathcal{N} (y; \theta,\sigma^2)$ with known $\sigma^2$. Our parameter of interest is defined as
\begin{equation}
\begin{aligned}
\theta_0(F_0) &= \argmax_{\theta} \int \log f_\theta(y) dF_0(y)\\
&= \argmin_{\theta} \int (y-\theta)^2 dF_0(y).
\end{aligned}
\end{equation}
If we set the derivative of the objective to 0, we obtain
\begin{equation}
\theta_0(F_0) = \int y dF_0(y).
\end{equation}

If we believe our parametric model to be accurate, we can place a  prior $\pi(\theta) =\mathcal{N}(\theta; 0,\tau^2)$ on $\theta$. The centering measure on our DP is thus
\begin{equation}
f_\pi(y) = \int f_\theta(y) d\pi(\theta)=\mathcal{N}(y; 0,\sigma^2+\tau^2).
\end{equation}
When $n=0$, our NPL prior $\tilde{\pi}(\theta)$ is approximately normal \cite{Yamato1984} from the properties of the DP:
\begin{equation}
\tilde{\pi}(\theta) \approx \mathcal{N}\left(\theta; 0,\frac{\sigma^2 + \tau^2}{1+\alpha}\right).
\end{equation}

 \subsubsection{Implementation and Results}
\begin{figure*}[!ht]
\begin{center}
\centerline{\includegraphics[width=\columnwidth]{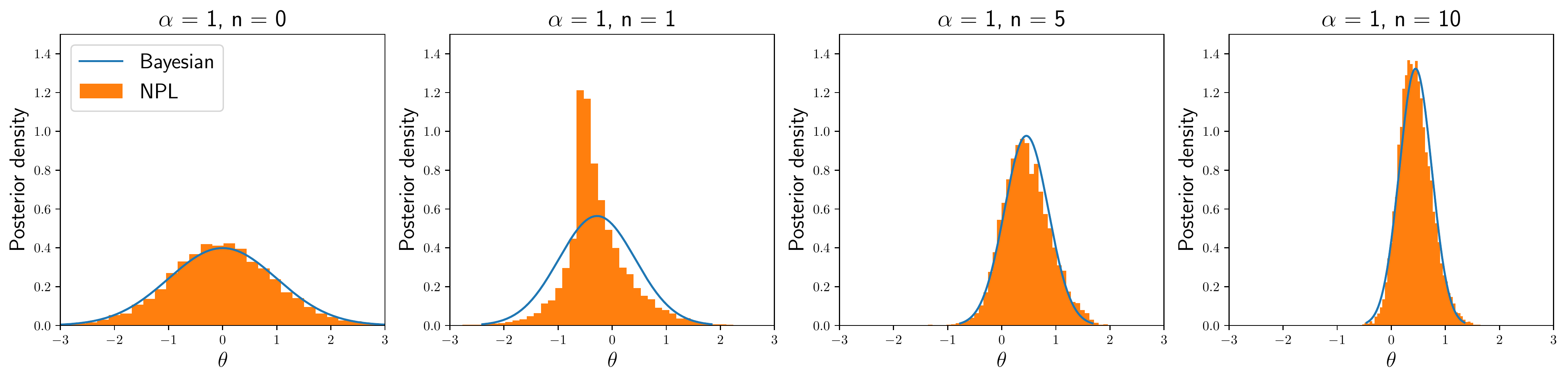}}
\vskip -0.1in
\caption{NPL posterior and Bayesian posterior for fixed $\alpha$ and increasing $n$ in normal location model}
\label{normn}
\end{center}
\vskip -0.3in
\end{figure*}

 \vspace{-0.05in}
We sample the observables $y \sim \mathcal{N}(1,1^2)$  and set our parametric prior variance to $\tau^2 = 1$. We simulate the NPL posterior in Figure \ref{normn} for various values of $n$ and $\alpha= 1$, and compare it to the tractable traditional Bayesian posterior with the same model $\{f_\theta, \pi(\theta) \}$.  For the NPL posterior bootstrap sampler, we generate $B=10000$ samples and truncate the DP at $T = 1000$.

\begin{figure*}[!ht]
\begin{center}
\centerline{\includegraphics[width=\columnwidth]{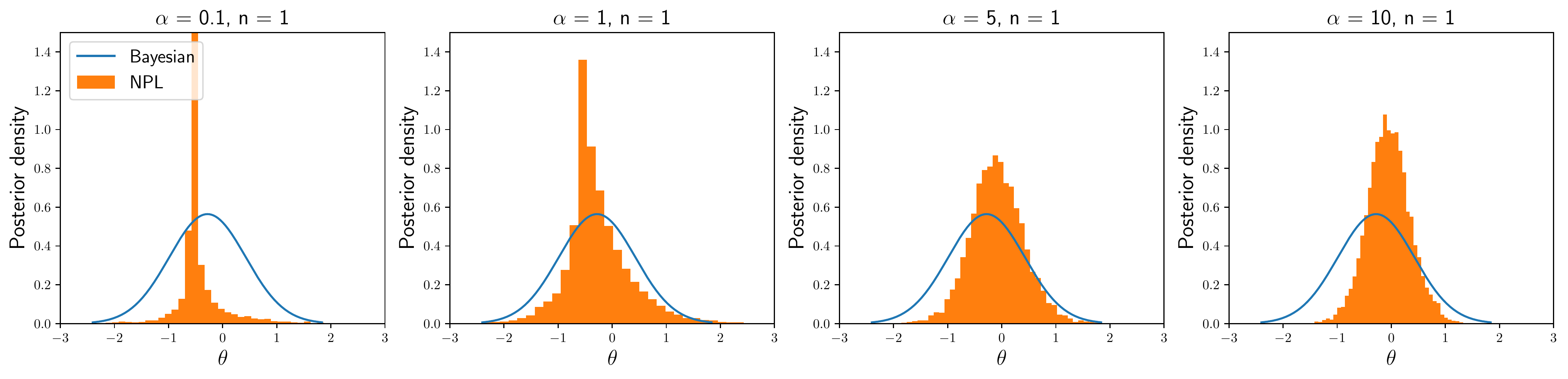}}
\caption{NPL posterior and Bayesian posterior for increasing $\alpha$ and fixed $n$ in normal location model}
\label{normalph}
\end{center}
\end{figure*}

We see from Figure \ref{normn} that the NPL prior is approximately normal ($n =0$), with same mean and variance due to the choice of $\alpha$. For large $n$, the NPL posterior and Bayesian posterior are similar, due to the first order correctness of the weighted likelihood bootstrap \cite{Newton1994}. For smaller values of $n$, the NPL posterior is non-normal, as our prior is not a conjugate prior on $\theta$.  For $n=1$, the sample observed is close to $0$ so the posterior uncertainty is small despite only observing one sample; this suggests that NPL may be better suited to moderate to large values of $n$.
Figure \ref{normalph} shows the effect on the NPL posterior of increasing prior strength $\alpha$ for $n=1$, which regularizes the posterior but also causes it to concentrate about $0$.

\subsection{Gaussian Mixture Model}

\subsubsection{Optimization Details}\label{weighted_EM}
We derive the EM algorithm that maximizes the weighted likelihood of the diagonal-covariance GMM:
\begin{equation*}
\begin{aligned}
\mathcal{L}^w(\theta) &= \sum_{i=1}^n w_i \log f_\theta (\mathbf{y}_i)\\ &= \sum_{i=1}^n w_i \left(\log f_\theta (\mathbf{y}_i,z_i) 
- \log f_\theta (z_i| \mathbf{y}_i)\right).
\end{aligned}
\end{equation*}
Taking an expectation over the posterior $f_{\theta'} ( z_{1:n} | \mathbf{y}_{1:n})$, we obtain
\begin{equation*}
\begin{aligned}
\mathcal{L}^w(\theta)  &= \sum_{i=1}^n w_i \sum_{z_i}  f_{\theta'} ( z_{i} | \mathbf{y}_{i}) \left(\log f_\theta (\mathbf{y}_i,z_i)  \right) - \sum_{i=1}^n w_i  \sum_{z_i}f_{\theta'} ( z_{i} | \mathbf{y}_{i}) \left(\log f_\theta (z_i| \mathbf{y}_i)\right) \\ &= \sum_{i=1}^n w_i Q^i (\theta | \theta') - \sum_{i=1}^n w_i H^i (\theta| \theta').
\end{aligned}
\end{equation*}

Taking the difference of the weighted likelihood with $\theta'$
\begin{equation*}
\begin{aligned}
\mathcal{L}^w(\theta) - \mathcal{L}^w(\theta') &= \sum_{i=1}^n w_i \left(Q^i(\theta|\theta') - Q^i(\theta'|\theta')\right)+ \sum_{i=1}^n w_i \left(H^i(\theta'|\theta') -H^i(\theta|\theta')\right).
\end{aligned}
\end{equation*}
From Gibbs' inequality,
\begin{equation*}
\begin{aligned}
H^i(\theta'|\theta') \geq H^i (\theta|\theta').
\end{aligned}
\end{equation*}
As all $w_i \geq  0$,
\begin{equation*}
\begin{aligned}
\mathcal{L}^w(\theta) - \mathcal{L}^w(\theta') \geq \sum_{i=1}^n w_i \left(Q^i(\theta|\theta') - Q^i(\theta'|\theta')\right).\\
\end{aligned}
\end{equation*}
So by maximizing $\sum_{i=1}^n w_i Q^i (\theta|\theta')$ w.r.t. $\theta$, we cannot decrease the weighted log-likelihood. As a reminder, the log-likelihood for each datapoint is
\begin{equation}
\log f_\theta (\mathbf{y}_i) =  \log \left(\sum_{k=1}^K \pi_k \mathcal{N}\left(\mathbf{y}_i; \boldsymbol{\mu}_k, \text{diag}\left(\boldsymbol{\sigma}^2_k\right) \right)\right).
\end{equation}
At the expectation step, we calculate
\begin{equation}
f_{\theta}(z_i = k | \mathbf{y}_i) = \frac{\prod_{j=1}^d \mathcal{N}(y_{ij}; \mu_{kj}, \sigma^2_{kj}) \pi_{k}}{\sum_{k=1}^K \prod_{j=1}^d \mathcal{N}(y_{ij}; \mu_{kj}, \sigma^2_{kj}) \pi_{k}}.
\end{equation}
The maximization step is then:
\begin{equation}
\begin{aligned}
\hat{\pi}_{k}&= \sum_{i=1}^n w_i f_{\theta'}(z_i = k | \mathbf{y}_i),\\
 \hat{\mu}_{kj}\ &= \frac{\sum_{i=1}^n w_i f_{\theta'}(z_i = k | \mathbf{y}_i) y_{ij}}{\hat{\pi}_{k}},\\
\hat{\sigma}^2_{kj} &= \frac{\sum_{i=1}^n w_i f_{\theta'}(z_i = k | \mathbf{y}_i) (y_{ij} - \hat{\mu}_{kj})^2}{\hat{\pi}_{k}}.
\end{aligned}
\end{equation}
\newpage
\subsubsection{Toy Example}\label{more_GMM}

\begin{figure*}[!ht]
\begin{center}
\centerline{\includegraphics[width=\columnwidth]{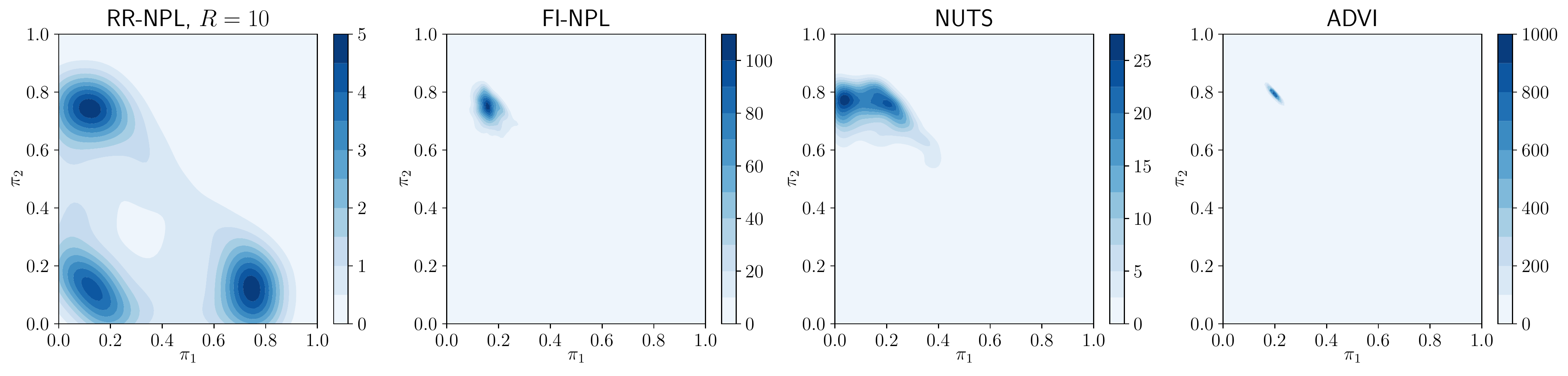}}
\caption{Posterior KDE of $(\pi_1,\pi_2)$ in $K=3$ toy separable GMM problem}
\label{pi}
\vskip 0.2in
\end{center}

\begin{center}
\centerline{\includegraphics[width=\columnwidth]{mu_sep.pdf}}
\caption{Posterior KDE of $(\mu_1,\mu_2)$ in $K=3$ toy separable GMM problem}
\label{mu_sup}
\end{center}

\vskip 0.2in
\begin{center}
\centerline{\includegraphics[width=\columnwidth]{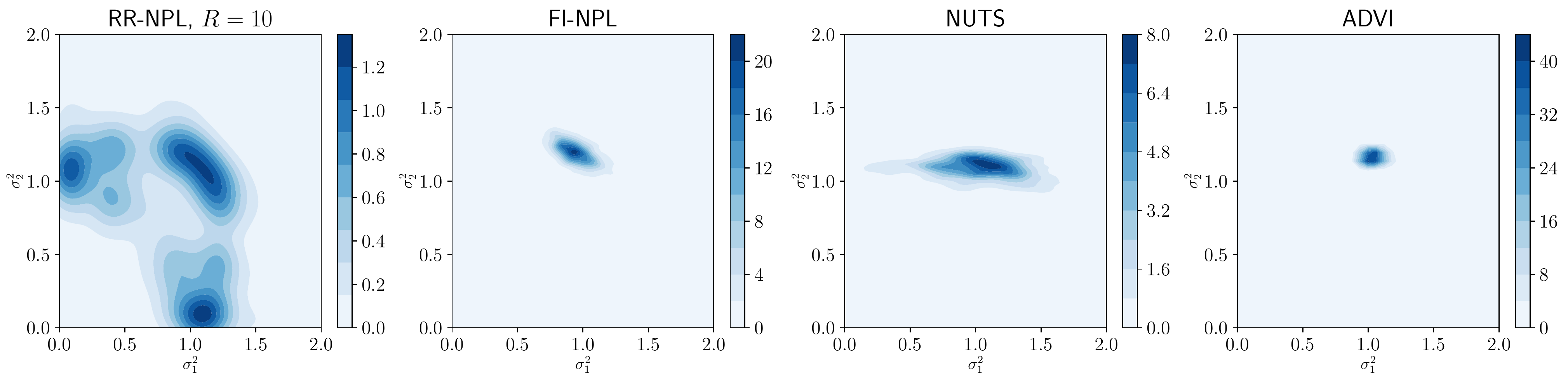}}
\caption{Posterior KDE of $(\sigma^2_1,\sigma^2_2)$ in $K=3$ separable toy GMM problem}
\label{sig}
\end{center}
\vskip -0.1in
\end{figure*}

\begin{figure*}

\vskip 0.2in
\begin{center}
\centerline{\includegraphics[width=\columnwidth]{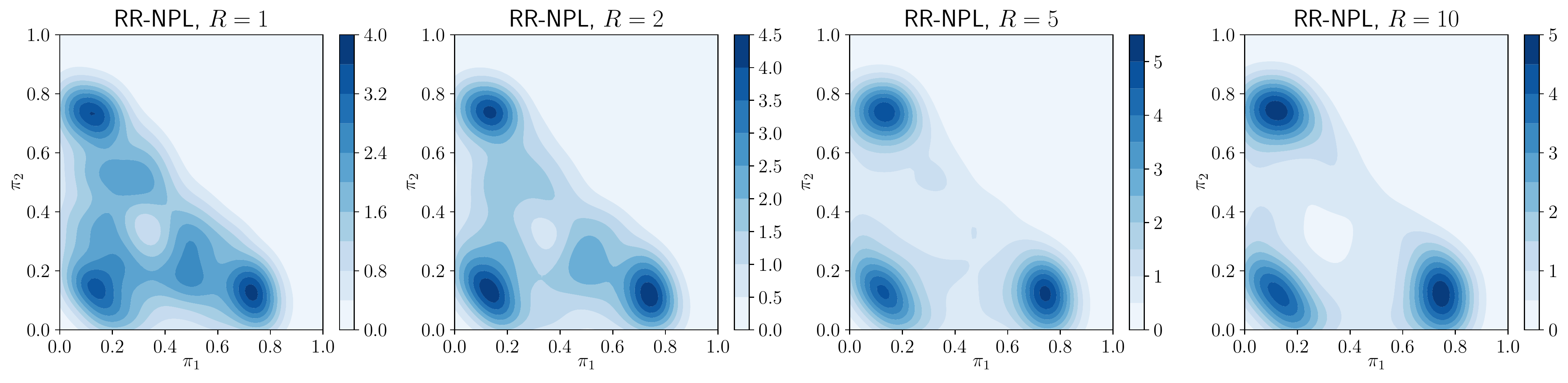}}
\caption{Posterior KDE of $(\pi_1,\pi_2)$ in $K=3$  separable toy GMM problem for increasing $R$}
\label{pi_reps}
\vskip 0.2in
\end{center}

\begin{center}
\centerline{\includegraphics[width=\columnwidth]{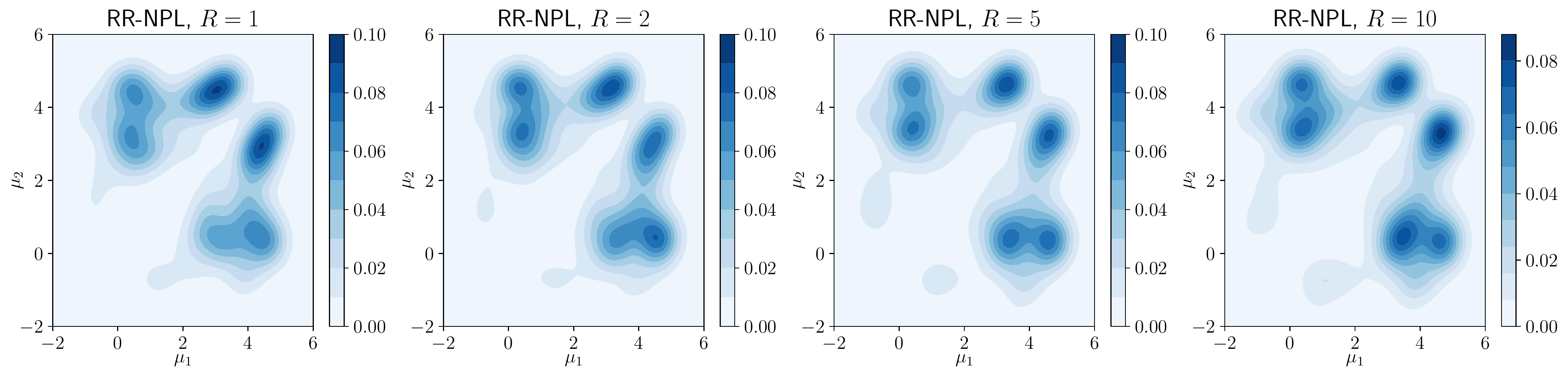}}
\vskip -0.1in
\caption{Posterior KDE of $(\mu_1,\mu_2)$ in $K=3$ toy GMM problem for RR-NPL with increasing $R$}
\label{mu_reps}
\vskip 0.2in
\end{center}

\begin{center}
\centerline{\includegraphics[width=\columnwidth]{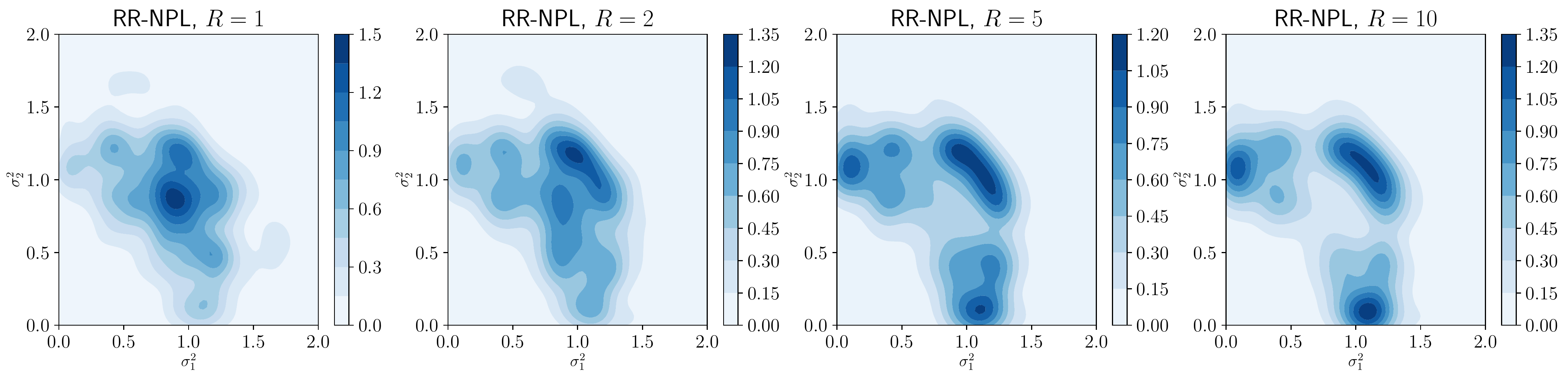}}
\caption{Posterior KDE of $(\sigma^2_1,\sigma^2_2)$ in $K=3$ separable  GMM toy problem for increasing $R$}
\label{sig_reps}
\end{center}
\end{figure*}

We see the posterior KDE plots for $(\pi_1,\pi_2)$, $(\mu_1,\mu_2)$ and  $(\sigma_1^2,\sigma_2^2)$ in Figures \ref{pi}, \ref{mu_sup}, \ref{sig}, and for increasing $R$ in Figures \ref{pi_reps}, \ref{mu_reps}, \ref{sig_reps}. For RR-NPL, we observe multimodality in addition to symmetry about the diagonal due to label-switching. Smaller values of $R$ exhibit an over-representation of local modes/saddles, and the posterior accuracy increases for larger $R$. We also show the run-times for different $R$ for RR-NPL in Table \ref{RRruntime}, and we see that the run-time increases roughly linearly with $R$.
\vskip -0.2in
\begin{table}[H]
\caption{Run-time (seconds) for 2000 posterior samples on Azure for different values of $R$ with RR-NPL}
\label{RRruntime}
\vskip 0.1in
\begin{center}
\begin{scriptsize}
\begin{sc}
\begin{tabular}{c||c|c|c|c}
  & $R=1$ &$R=2$ &$R=5$ & $R =10$   \\
\hline \hline
 Toy Sep & 4.9 $\pm$ 0.6 &8.0 $\pm$1.1& 19.0$\pm$ 2.1   &  37.2s $\pm$ 4.5s   \\
\end{tabular}
\end{sc}
\end{scriptsize}
\end{center}
\end{table}

\subsubsection{Comparison to Importance Sampling}\label{Imp}
As suggested by our helpful reviewers and meta-reviewer, we include a discussion and empirical comparison with importance sampling (IS) here. The efficacy of IS hinges on finding a good approximating proposal density, which is in itself a challenging research question. Generic proposals can lead to very large and difficult to detect errors (e.g. see \citet[pg. 534]{Bishop:2006:PRM:1162264}). Moreover, the variance of the IS approximation is driven by the variance of the importance weights $p(x)/q(x)$, for $x\sim q(x)$ approximating $p(x)$. The proposal distribution needs to capture all aspects of the target distribution and not just the modes, otherwise the ratio $p(x)/q(x)$ may not be bounded. This makes IS challenging to apply in moderate to high dimensional problems and especially when there is multimodality; we will now demonstrate this in our toy GMM example from Section \ref{GMMsec} of the main paper.

We set the task of estimating the mean log pointwise predictive density (LPPD) \cite{gelman2013bayesian} of the 250 held-out test data points; the LPPD is defined in Section \ref{pred_perf}. We use self-normalized importance sampling (SNIS) as the posterior is unnormalized, so the estimate of the posterior predictive is defined as
\begin{equation}
p(\tilde{y}|y_{1:n}) \approx \frac{\sum_{b=1}^B w_b f(\tilde{y} | \boldsymbol{\theta}_b) }{\sum_{b=1}^B w_b} = \sum_{b=1}^B \tilde{w}_b f(\tilde{y} | \boldsymbol{\theta}_b), \quad w_b = \frac{f(y_{1:n}|\boldsymbol{\theta}_b)\pi(\boldsymbol{\theta}_b)}{q(\boldsymbol{\theta}_b)}, \quad \boldsymbol{\theta}_b \sim q(\boldsymbol{\theta})
\end{equation}
where $\boldsymbol{\theta} = \{ \boldsymbol{\pi}, \boldsymbol{\mu}, \boldsymbol{\sigma}\}$ is $9$-dimensional, and $\{f(y|\boldsymbol{\theta}), \pi(\boldsymbol{\theta})\}$ is the GMM likelihood and prior as defined in (\ref{GMM}) of the main paper. The choice of the proposal is non-trivial, as both the posterior and the function being integrated are multimodal. We use a broad proposal of the same form as the prior as shown below:
\begin{equation}
q(\boldsymbol{\pi}) = \text{Dir}(0.1,\dots,0.1), \quad q(\mu_{kj}) = \mathcal{N}(0,25), \quad q(\sigma_{kj} ) = \text{logNormal}(0,1).
\end{equation}
The proposal has support in the true value of $\boldsymbol{\theta}$ and thus should have support in regions of high $f(\tilde{y} |  \boldsymbol{\theta})$. As we are integrating over $n_\text{test}$ distinct likelihoods with varying multimodality, a broad proposal is appropriate. We carry out SNIS with $B=10^7$ implemented on the same virtual machines, where this choice of $B$ is determined by approximately matching the time required to produce $2000$ posterior samples with NPL. We repeat 10 runs with the same training and test set but vary the seed for the samplers, and report the mean and standard error (SE) of the mean LPPD. For SNIS, we also report the effective sample size, defined as $\text{ESS}=  1/\sum_{b=1}^B \tilde{w}_b^2$. This is shown below in Table \ref{IS}, and we see that the SE for SNIS is an order of magnitude larger than that of RR-NPL. Furthermore, the ESS of SNIS is extremely poor, likely due to the difficulty in selecting a good proposal in this 9-dimensional problem. We notice that most of the weights are very close to 0 except for a few that dominate. These effects will be increasingly amplified in higher dimensions, and is thus why IS fails in problems of even moderate dimensionality.

\vskip -0.2in
\begin{table}[H]
\caption{Performance on held-out test data for toy GMM}
\label{IS}
\vskip 0.1in
\begin{center}
\begin{scriptsize}
\begin{sc}
\begin{tabular}{c||c|c}
  &RR-NPL& SNIS    \\
\hline \hline
Mean of LPPD estimate &-{1.7984 }& -1.8070 \\ 
SE of LPPD estimate & $2 \times 10^{-4}$&$4.4 \times 10^{-3}$\\
ESS & $2000$& $1.75\pm 0.80$ \\
Run-time &29.3s $\pm$ 0.5s &31.0s $\pm$ 8.7s
\end{tabular}
\end{sc}
\end{scriptsize}
\end{center}
\end{table}
\vskip -0.2in
\subsubsection{Comparison to NPL with Mixture of Dirichlet Processes Prior}\label{MDP}

As explained in the main paper, NPL with a mixture of DPs prior (MDP-NPL) as introduced in \citet{Lyddon} requires accurate sampling of the Bayesian posterior, which MCMC and VB may not be able to provide. Another difference is the meaning of the parameter $\alpha$ in the two NPL schemes, which is the concentration of the MDP and the DP. In MDP-NPL, the concentration represents the strength of belief that the centering traditional Bayesian model is correctly specified, whereas in NPL with a DP prior (DP-NPL) it is the strength of belief that $F_\pi$ is the true sampling distribution. We see in Section \ref{prior_intuit} that the limit of $\alpha \to \infty$ gives different results between the two NPL schemes, while $\alpha \to 0$ gives the same limit. 

\begin{figure*}[!ht]
\begin{center}
\centerline{\includegraphics[width=\columnwidth]{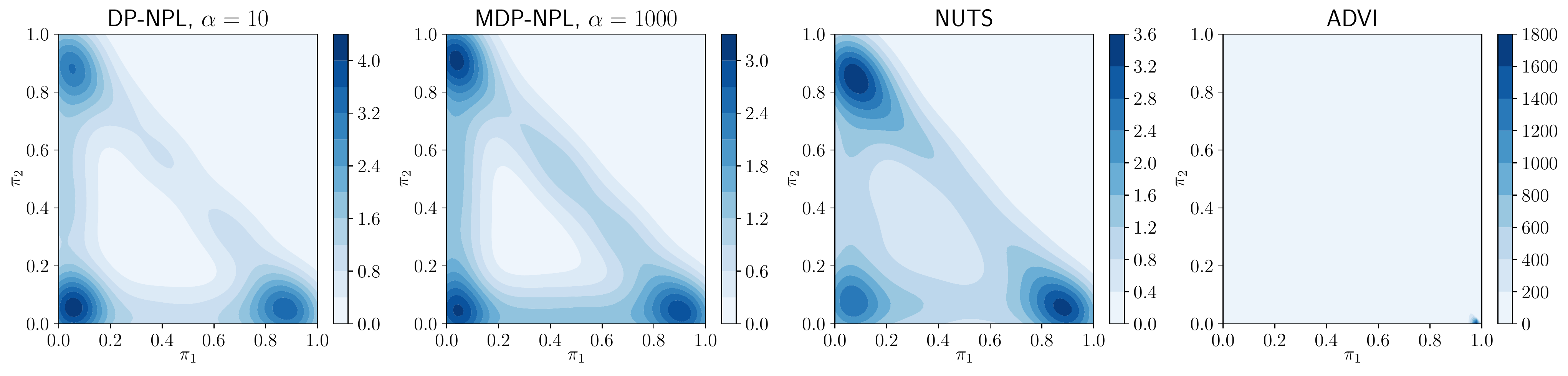}}
\caption{Posterior KDE of $(\pi_1,\pi_2)$ in $K=3$ inseparable toy GMM problem 2}
\label{pi_insep}
\vskip 0.2in
\end{center}
\begin{center}
\centerline{\includegraphics[width=\columnwidth]{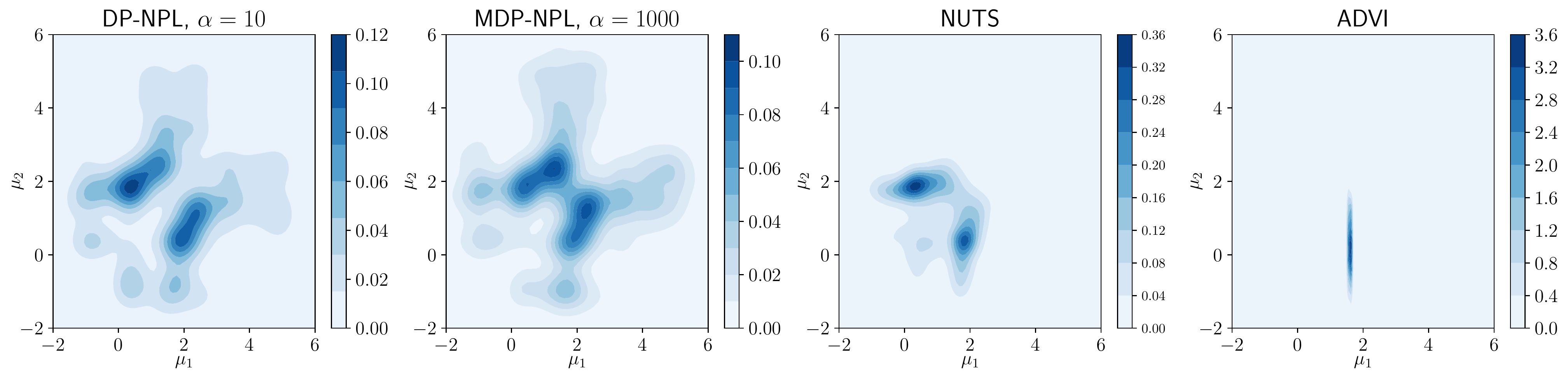}}
\caption{Posterior KDE of $(\mu_1,\mu_2)$ in $K=3$ inseparable toy GMM problem}
\label{mu_insep}
\end{center}

\begin{center}
\vskip 0.2in
\centerline{\includegraphics[width=\columnwidth]{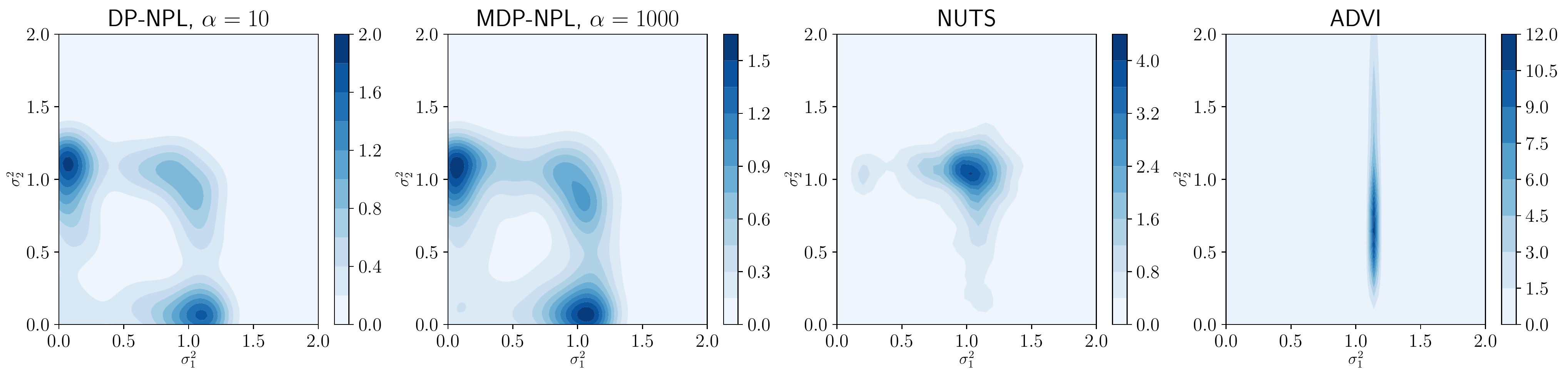}}
\caption{Posterior KDE of $(\sigma^2_1,\sigma^2_2)$ in $K=3$ inseparable toy GMM problem}
\label{sig_insep}
\end{center}
\vskip -0.4in
\end{figure*}

As evident in Figure \ref{mu} of the main paper, NUTS and ADVI clearly fail at representing the multimodality in our toy GMM problem, and as a result it is not possible to carry out MDP-NPL in that example. We thus compare DP-NPL to MDP-NPL experimentally in an easier toy GMM problem in which NUTS can represent the posterior accurately. We carry out the same experiment with alternative GMM parameters:
\begin{equation*}
\begin{aligned}
\boldsymbol{\pi}_0 = \{0.1,0.3,0.6\}, \hspace{2mm} \boldsymbol{\mu}_0 = \{0,1,2\}, \hspace{2mm} \boldsymbol{\sigma}_0^2 = \{1,1,1\}.
\end{aligned}
\end{equation*}
The means are closer together, and so NUTS can mix properly as shown in  Figure \ref{pi_insep}. For MDP-NPL, we center the MDP with the Bayesian model given in (\ref{GMM}) of the main paper, set $\alpha = 1000$ and carry out the posterior bootstrap step using the posterior samples generated by NUTS and maximizing the weighted likelihood. For DP-NPL, we elicit the centering measure $f_\pi(y) = \mathcal{N}(y; 0,1)$ and set $\alpha = 10$. For both NPL schemes, we carry out 10 random restarts for each posterior sample with the same initialization scheme as in the main paper. 

The posterior KDE plots for $(\pi_1,\pi_2)$, $(\mu_1,\mu_2)$ and  $(\sigma_1^2,\sigma_2^2)$ are shown in in Figures \ref{pi_insep}, \ref{mu_insep}, \ref{sig_insep}. The difference between the DP-NPL and MDP-NPL posteriors is small,  and both can represent the multimodality well. Predictively, DP-NPL and MDP-NPL perform similarly as shown in Table \ref{GMMinsep}, but the run-times are much greater for MDP-NPL as shown in Table \ref{GMMinsepruntime}. This is because we first need to generate the Bayesian posterior samples via NUTS before we can proceed to the posterior bootstrap, and so the run-time of NUTS is still the bottleneck.

\begin{table}[H]
\caption{Mean LPPD on held-out test data for inseparable GMM}
\vskip 0.05in
\label{GMMinsep}
\begin{center}
\begin{scriptsize}
\begin{sc}
\begin{tabular}{c||c|c|c|c}
  &DP-NPL& MDP-NPL &NUTS &ADVI   \\
\hline \hline
Toy &-1.612 $\pm$ 0.038& -1.610 $\pm$ 0.037 &  -1.609$\pm$0.037 &-1.613 $\pm$ 0.038\\
\end{tabular}
\end{sc}
\end{scriptsize}
\end{center}
\end{table}
\vskip -0.1in
\begin{table}[H]
\caption{Run-time for 2000 samples for inseparable GMM}
\label{GMMinsepruntime}
\vskip 0.05in
\begin{center}
\begin{scriptsize}
\begin{sc}
\begin{tabular}{c||c|c|c|c}
  & DP-NPL& MDP-NPL &NUTS&ADVI   \\
\hline \hline
 Toy&52.6s$ \pm$ 6.3s &3m2s $\pm$ 13s &2m12s $\pm$ 12s  &  0.8s $\pm$ 0.1s     \\
\end{tabular}
\end{sc}
\end{scriptsize}
\end{center}
\end{table}

\subsection{Logistic Regression with Automatic Relevance Determination Priors}

\subsubsection{Predictive Performance}\label{pred_perf}

For all the following measures of predictive performance on held-out test data, we can use a Monte Carlo estimate of the predictive distribution of a test data point $(\tilde{y},\tilde{x})$:
\begin{equation}
\begin{aligned}
p(\tilde{y} |\tilde{x},y_{1:n},x_{1:n}) &=\int f(\tilde{y} | \tilde{x}, \boldsymbol{\beta}) d\tilde{\pi}(\boldsymbol{\beta}|y_{1:n},x_{1:n}) \\ &\approx \frac{1}{B}\sum_{b=1}^B f(\tilde{y}|\tilde{x},\boldsymbol{{\beta}}_b), \\ \boldsymbol{\beta}_b &\sim \tilde{\pi}(\boldsymbol{\beta}|y_{1:n},x_{1:n}),
\end{aligned}
\end{equation}
where $f(y| x, \boldsymbol{\beta})$ is the likelihood, $\tilde{\pi}(\boldsymbol{\beta}|y_{1:n},x_{1:n})$  is the NPL or Bayesian posterior, $B$ is the number of posterior samples, and $(y_{1:n},x_{1:n})$ is the training set. We evaluate the mean LPPD of held-out test data as a measure of predictive performance:
\begin{equation}
\begin{aligned} 
\text{Mean LPPD} =\frac{1}{n_{\text{test}}}\sum_{i=1}^{n_{\text{test}}} \log p(\tilde{y}_i |\tilde{x}_i,y_{1:n},x_{1:n}).
\end{aligned}
\end{equation}

Below, we additionally include the mean squared error (MSE) here on held-out test data, defined
\begin{equation} \text{MSE} = \frac{1}{n_{\text{test}}} \sum_{i=1}^{n_{\text{test}}} ( p(\tilde{y}_i |\tilde{x}_i,y_{1:n},x_{1:n}) - \tilde{y}_i)^2.
\end{equation}

Finally, we also report the percentage accuracy, defined
\begin{equation}
\begin{aligned}
\text{P.a.} &= \frac{1}{n_{\text{test}}} \sum_{i=1}^{n_{\text{test}}} \hat{y}_i^{\tilde{y}_i}  (1-\hat{y}_i)^{(1-{\tilde{y}_i})}, \\ \hat{y}_i&= \mathbb{I}(p(\tilde{y}_i |\tilde{x}_i,y_{1:n},x_{1:n}) >0.5)
\end{aligned}
\end{equation}
where $\mathbb{I}$ is the indicator function. 
\subsubsection{Sparsity Measure}\label{sparsity_meas}
For the sparsity results, we simply calculate the posterior mean $\hat{\boldsymbol{\beta}} = \frac{1}{B} \sum_{b=1}^B \boldsymbol{\beta}_b$, where $ \boldsymbol{\beta}_b \sim \tilde{\pi}(\boldsymbol{\beta}|y_{1:n},x_{1:n})$ as above. We then report the percentage of components of $\hat{\boldsymbol{\beta}}$ that have absolute value less than $\epsilon$.
\subsubsection{Optimization Details}\label{loss_grad}

L-BFGS-B \cite{Zhu1997} is a quasi-Newton method which requires the gradient, which for the marginal Student-t distribution is defined for $j\in \{1,\dots,d\}$ as
\begin{equation}
\begin{aligned}
\frac{\partial l(y,x,\boldsymbol{\beta},\beta_0)}{\partial \beta_j }&= -\left(y - \eta \right)x_{j}  + \gamma\left(\frac{2a+1}{2b + \beta_j^2}\right)\beta_j, \\
\frac{\partial l(y,x,\boldsymbol{\beta},\beta_0)}{\partial \beta_0 }&= -\left(y - \eta \right).
\end{aligned}
\end{equation}

\subsubsection{Additional Results}\label{more_logreg}

\begin{figure*}[!ht]
\vskip -0.2in
\begin{center}
\centerline{\includegraphics[width=0.6\columnwidth]{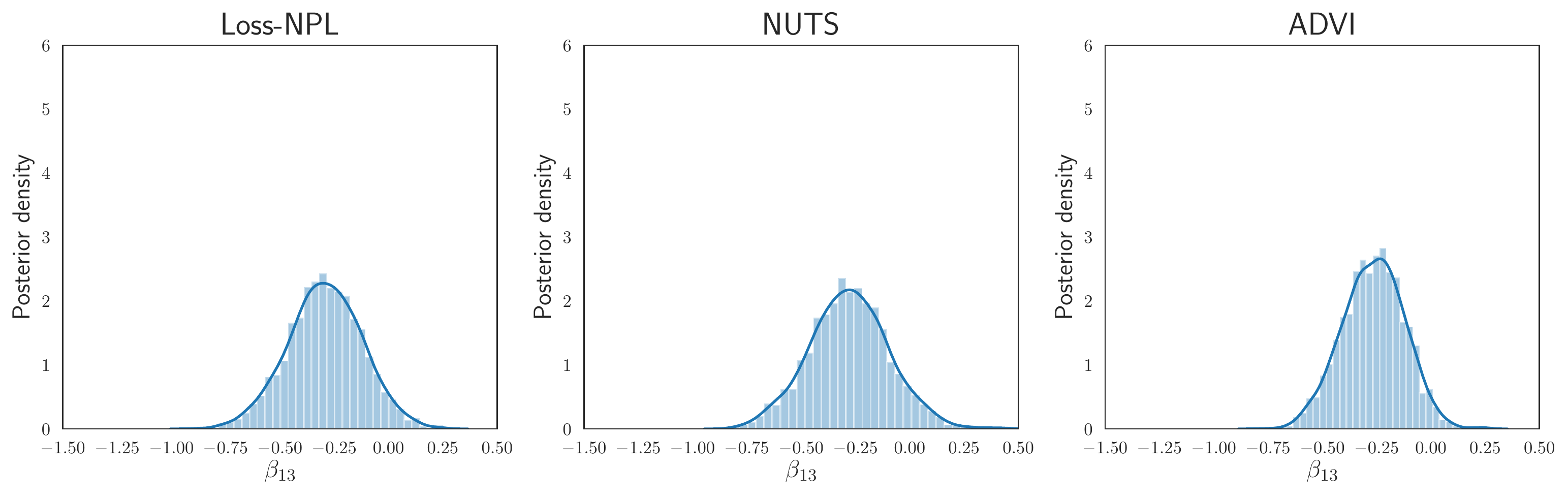}}
\caption{Posterior marginal KDE of $\beta_{13}$ for `Adult' dataset}
\label{betaadult}
\end{center}
\vskip -0.2in
\end{figure*}

We can see in Tables \ref{MSE}, \ref{PA} that loss-NPL performs equally or better than NUTS and ADVI predictively in MSE and classification accuracy as well as LPPD. A posterior marginal density plot for $\beta_{13}$ in the `Adult' dataset is shown in Figure \ref{betaadult} for reference. 
 \vskip -0.1in
\begin{table}[H]
\caption{MSE on held-out test data}
\label{MSE}
\vskip 0.15in
\begin{center}
\begin{scriptsize}
\begin{sc}
\begin{tabular}{c||c|c|c}
 Data Set & Loss-NPL & NUTS &ADVI  \\
\hline \hline
 Adult & \textbf{0.104} $\pm$ 0.001&   \textbf{0.104 }$\pm$ 0.001 & 0.105$\pm$  0.002\\
 Polish & \textbf{0.056} $\pm$ 0.011&  0.524 $\pm$0.236 &0.058 $\pm$  0.011\\
 Arcene &\textbf{0.134}$\pm$0.026&0.152$\pm$ 0.014&0.143 $\pm$0.020	
\end{tabular}
\end{sc}
\end{scriptsize}
\end{center}
\vskip -0.4in
\end{table}

\begin{table}[H]
\caption{Predictive accuracy \%  on held-out test data}
\label{PA}
\vskip 0.15in
\begin{center}
\begin{scriptsize}
\begin{sc}
\begin{tabular}{c||c|c|c}
 Data Set & Loss-NPL & NUTS &ADVI  \\
\hline \hline
 Adult & \textbf{84.92} $\pm$ 0.29 &\textbf{84.92}$\pm$ 0.30 &  84.84 $\pm$ 0.30  \\
 Polish & \textbf{93.65 }$\pm$ 1.68 & 37.27$\pm$ 23.81 &93.51$\pm$ 1.56 \\
 Arcene &\textbf{81.73} $\pm$3.79 &77.80$\pm$4.22 & 79.70$\pm$ 3.40
\end{tabular}
\end{sc}
\end{scriptsize}
\end{center}
\end{table}

\subsection{Bayesian Sparsity-path-analysis}
\subsubsection{Values of $\boldsymbol{\beta}$}\label{beta_vals}
We follow the setting of $\boldsymbol{\beta}$ in \citet{Lee2012BayesianSO}: the 5 non-zero indices and their respective values are
 \begin{equation*}
 \begin{aligned}
 \mathcal{I}&=\{10, 14, 24, 31, 37 \}, \\
 \beta_{\mathcal{I}}&= \{-0.2538, 0.4578, -0.1873, -0.1498, 0.0996 \}.
 \end{aligned}
 \end{equation*}
\subsubsection{Variable selection}\label{more_gen}
\begin{figure}[ht]
\vskip 0.05in
\begin{center}
\centerline{\includegraphics[width=0.5\columnwidth]{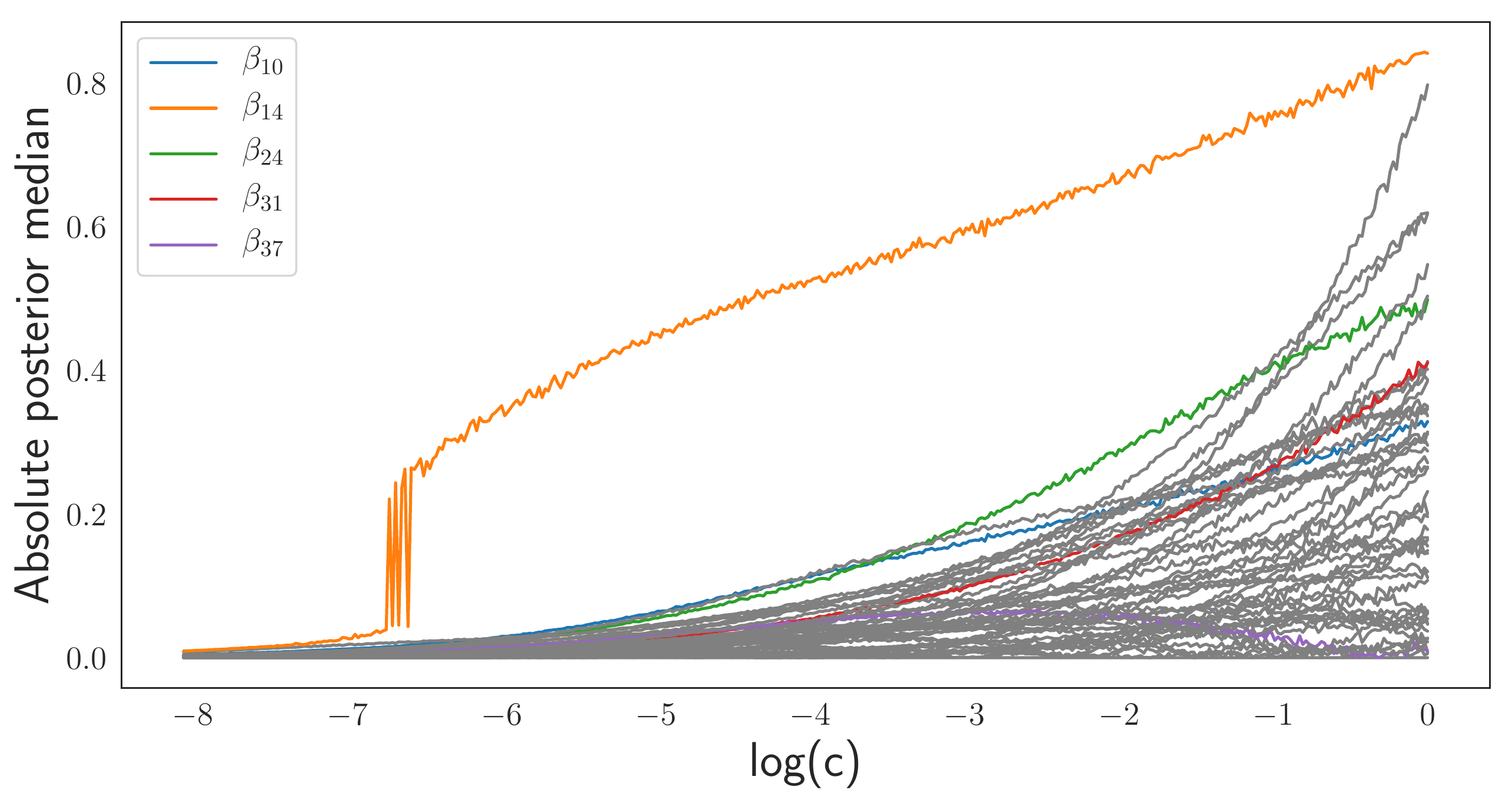}}
\vskip -0.05in
\caption{Lasso-type plot for absolute posterior medians of $\boldsymbol{\beta}$ against $\log(c)$  from genetic dataset with non-zero components in colour}
\label{sparseabsmedian}
\end{center}
\vskip -0.25in
\end{figure}

We see more clearly from Figure \ref{sparseabsmedian} that $\beta_{10}$, $\beta_{14}$ and $\beta_{24}$ have early predictive power, with one other null-coefficient showing early predictive importance. The other two non-zero coefficients are masked.

\end{document}